\documentclass[10pt,twocolumn,letterpaper]{article}

\usepackage{iccv}
\usepackage{times}
\usepackage{epsfig}
\usepackage{graphicx}
\usepackage{amsmath}
\usepackage{amssymb}
\usepackage{booktabs}       
\usepackage{multirow}   
\usepackage{makecell}
\usepackage[dvipsnames]{xcolor}
\usepackage{colortbl}
\usepackage{subcaption}
\usepackage{caption}
\usepackage{pifont}
\usepackage{lipsum}
\usepackage[percent]{overpic}
\usepackage{float}
\usepackage{pdfpages}
\usepackage[sort,nocompress]{cite}
\usepackage[accsupp]{axessibility}

\usepackage[pagebackref=true,breaklinks=true,letterpaper=true,colorlinks,bookmarks=false]{hyperref}

\newcommand{\mcc}[1]{\multicolumn{#1}{c}}
\newcommand{\mcp}[1]{\multicolumn{#1}{c@{\hspace{30pt}}}}
\definecolor{Gray}{gray}{0.90}
\newcolumntype{a}{>{\columncolor{Gray}}r}
\newcolumntype{b}{>{\columncolor{Gray}}c}
\newcommand{\B}[1]{\textcolor{blue}{\textbf{#1}}}

\newcommand{\cmark}{\ding{51}}%
\def\model{Episodic Transformer\xspace}
\def\modelshort{E.T.\xspace}

\newcommand{\alframe}[2]{
    \begin{overpic}[width=0.188\textwidth]{#1}
        \put (10,10) {\textcolor{white}{$t = {#2}$}}
    \end{overpic}
}

\iccvfinalcopy 

\ificcvfinal\fi

\begin{document}

\title{Episodic Transformer for Vision-and-Language Navigation} 

\author{
Alexander Pashevich$^{1}$\thanks{Work done as an intern at Google Research.}
\qquad
Cordelia Schmid$^{2}$
\qquad
Chen Sun$^{2,3}$
 \\
$^{1}$ Inria \quad $^{2}$ Google Research \quad $^{3}$ Brown University
}

\maketitle
\ificcvfinal\thispagestyle{empty}\fi

\begin{abstract}
Interaction and navigation defined by natural language instructions in dynamic environments pose significant challenges for neural agents. This paper focuses on addressing two challenges: handling long sequence of subtasks, and understanding complex human instructions. We propose \model (\modelshort), a multimodal transformer that encodes language inputs and the full episode history of visual observations and actions. 
To improve training, we leverage synthetic instructions as an intermediate representation that decouples  understanding the visual appearance of an environment from the variations of natural language instructions. We demonstrate that encoding the history with a transformer is critical to solve compositional tasks, and that pretraining and joint training with synthetic instructions further improve the performance. Our approach sets a new state of the art on the challenging ALFRED benchmark, achieving $38.4\%$ and $8.5\%$ task success rates on seen and unseen test splits.
\end{abstract}
\section{Introduction}

\begin{figure}
\centering
\includegraphics[width=\linewidth]{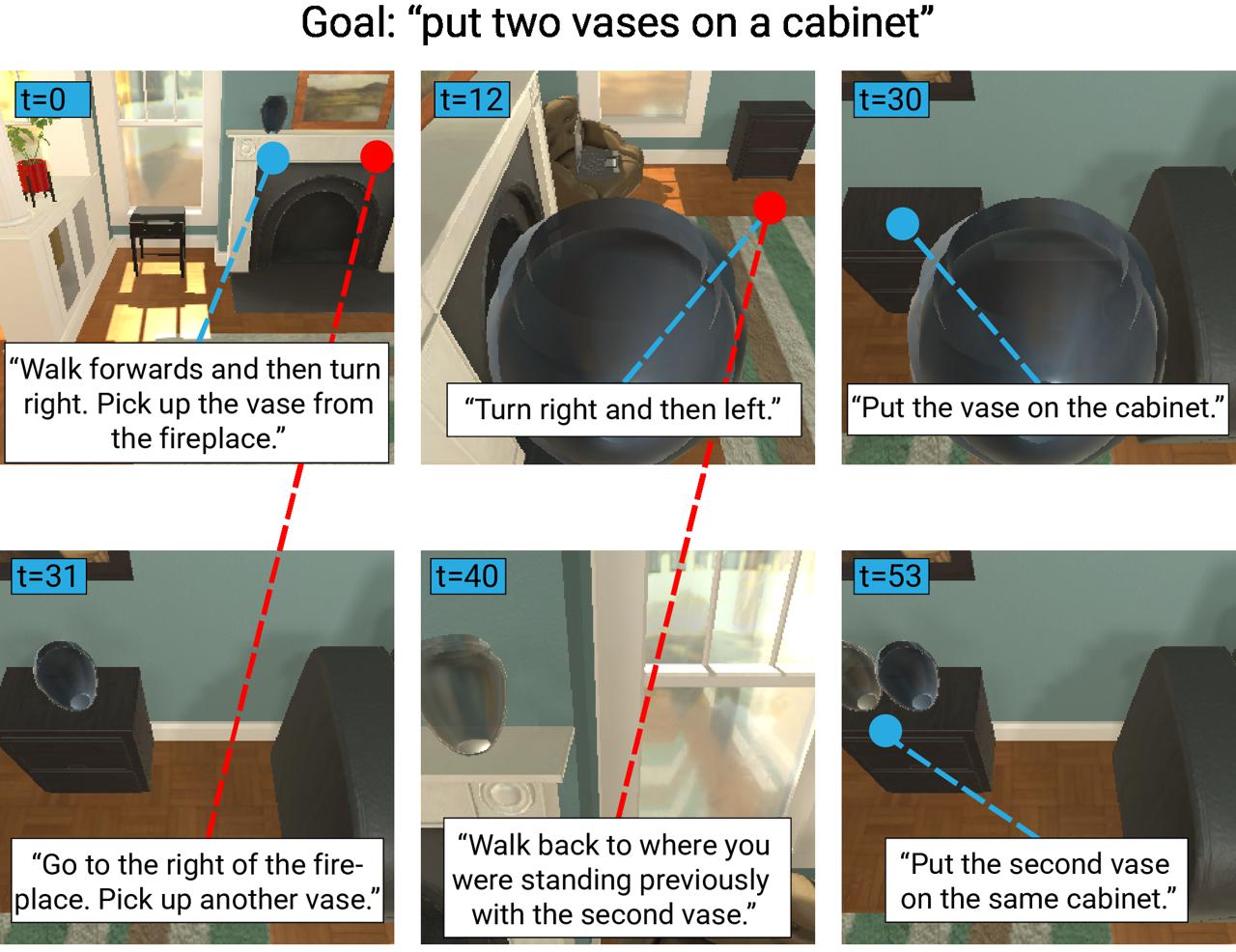}
\caption{
An example of a compositional task in the ALFRED dataset~\cite{ALFRED20} where the agent is asked to bring two vases to a cabinet.
We show several frames from an expert demonstration with corresponding step-by-step instructions.
The instructions expect the agent to be able to navigate to a \textit{fireplace} which is not visible in its current egocentric view and to remember its previous location by referring to it as \textit{"where you were standing previously"}. 
}
\vspace{-0.5cm}
\label{fig:teaser}
\end{figure}

Having an autonomous agent performing various household tasks is a long-standing goal of the research community. To benchmark research progress, several simulated environments~\cite{mattersim,ALFRED20, puig2018virtualhome} have recently emerged where the agents navigate and interact with the environment following natural language instructions. Solving the vision-and-language navigation (VLN) task requires the agent to ground human instructions in its embodied perception and action space. In practice, the agent is often required to perform long compositional tasks while observing only a small fraction of the environment from an egocentric point of view. Demonstrations manually annotated with human instructions are commonly used to teach an agent to accomplish specified tasks.

This paper attempts to address two main challenges of VLN: (1) handling highly compositional tasks consisting of many subtasks and actions; (2) understanding the complex human instructions that are used to specify a task. Figure~\ref{fig:teaser} shows an example task that illustrates both challenges. We show six key steps from a demonstration of 53 actions. To fulfill the task, the agent is expected to remember the location of a \textit{fireplace}  at $t=0$ and use this knowledge much later (at $t=31$). It also needs to solve object- (\eg ``another vase'') and location-grounded (\eg ``where you were standing previously'') coreference resolution in order to understand the human instructions.

Addressing the first challenge requires the agent to remember its past actions and observations. Most recent VLN approaches rely on recurrent architectures~\cite{tan-etal-2019-learning, rcm2019, Zhu_2020_CVPR, ma2019selfmonitoring} where the internal state is expected to keep information about previous actions and observations. However, the recurrent networks are known to be inefficient in capturing long-term dependencies~\cite{NIPS2015_277281aa} and may fail to execute long action sequences~\cite{graves2016hybrid, ALFRED20}.
Motivated by the success of the attention-based transformer architecture~\cite{attentionisall} at language understanding~\cite{devlin2018bert, brown2020language} and multimodal learning~\cite{Sun_2019_ICCV, object_attention2020}, we propose to use a transformer encoder to combine multimodal inputs including camera observations, language instructions, and previous actions. The transformer encoder has access to the history of the \textit{entire episode} to allow long-term memory and outputs the action to take next. We name our proposed architecture \textbf{Episodic Transformer (E.T.)}.

Addressing the second challenge requires revisiting different ways to specify a task for the autonomous agent. We observe that domain-specific language~\cite{pddl1998} and temporal logic~\cite{Gopalan-RSS-18, ltlbook1992} can unambiguously specify the target states and (optionally) their temporal dependencies, while being decoupled from the visual appearance of a certain environment and the variations of human instructions. We hypothesize that using these \textit{synthetic instructions} as an intermediate interface between the  human and the agent would help the model to learn more easily and generalize better. To this end, we propose to pretrain the transformer-based language encoder in E.T. by predicting the synthetic instructions from human instructions. We also explore joint training, where human instructions and synthetic instructions are mapped into a shared latent space.

To evaluate the performance of E.T., we use the ALFRED dataset~\cite{ALFRED20} which consists of longer episodes than the other vision-and-language navigation datasets~\cite{mattersim, puig2018virtualhome, touchdown} and also requires object interaction. We experimentally show that E.T. benefits from full episode memory and is better at solving tasks with long horizons than recurrent models. We also observe significant gains by pretraining the language encoder with the synthetic instructions. Furthermore, we show that when used for training jointly with natural language such intermediate representations outperform conventional data augmentation techniques for vision-and-language navigation~\cite{fried2018speaker} and work better than image-based annotations~\cite{lynch2019play}.

In summary, our two main contributions are as follows. First, we propose Episodic Transformer (E.T.), an attention-based architecture for vision-and-language navigation, and demonstrate its advantages over recurrent models.
Second, we propose to use synthetic instructions as the intermediate interface between the human and the agent.
Both contributions combined allow us to achieve a new state-of-the-art on the challenging ALFRED dataset.

Code and models are available on the project page\footnote{ \url{https://github.com/alexpashevich/E.T.}}.

\section{Related work}

\begin{figure*}
\centering
\includegraphics[width=\linewidth]{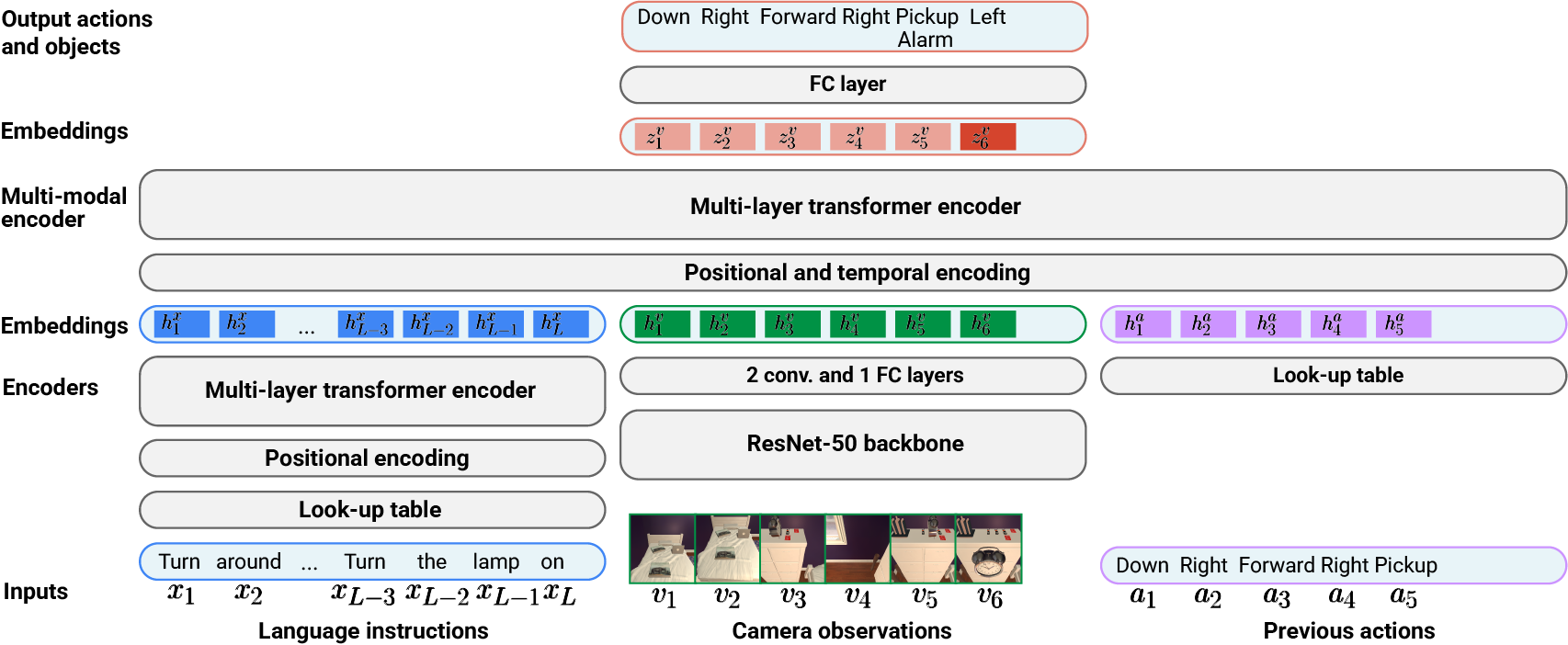}
\caption{
Episodic Transformer (E.T.) architecture.
To predict the next action, the E.T. model is given a natural language instruction $x_{1:L}$, visual observations since the beginning of an episode $v_{1:t}$, and previously taken actions $a_{1:t-1}$.
Here we show an example that corresponds to the $6^{\textrm{th}}$ timestep of an episode: $t = 6$. After processing $x_{1:L}$ with a transformer-based language encoder, embedding $v_{1:t}$ with a ResNet-50 backbone and passing $a_{1:t-1}$ through a look-up table, the agent outputs $t$ actions.
During training we use all predicted actions for a gradient descent step.
At test time, we apply the last action $a_t$ to the environment.
}
\vspace{-0.3cm}
\label{fig:archi}
\end{figure*}

\noindent \textbf{Instruction following agents.} Building systems to understand and execute human instructions has been the subject of many previous works~\cite{bugmann2001using,branavan2009reinforcement,tenorth2010understanding,macmahon2006walk,chen2011learning,bollini2013interpreting,misra2017mapping,misra2016tell,paul2018efficient,lynch2019play}. Instruction types include structured commands or logic programs~\cite{ltlbook1992,pddl1998,puig2018virtualhome}, natural language~\cite{chen2011learning,tellex2011understanding}, target state images~\cite{lynch2019play}, or a mix~\cite{lynch2020language}.
While earlier work focuses on mapping instructions and structured world states into actions~\cite{mei2016listen,andreas2015alignment,Patel2019LearningTG}, it is desirable for the agents to be able to handle raw sensory inputs, such as images or videos. To address this, the visual-and-language navigation (VLN) task is proposed to introduce rich and unstructured visual context for the agent to explore, perceive and execute upon~\cite{mattersim,ku2020room,touchdown,mehta2020retouchdown,krantz2020beyond}. The agent is requested to navigate to the target location based on human instructions and real, or photo-realistic image inputs, implemented as navigation graphs~\cite{mattersim,touchdown} or a continuous environment~\cite{krantz2020beyond} in simulators~\cite{todorov2012mujoco,ai2thor,robothor2020,savva2019habitat}. More recently, the ALFRED environment~\cite{ALFRED20} introduces the object interaction component to complement visual-language navigation. It is a more challenging setup as sequences are longer than in other vision-language navigation datasets and all steps of a sequence have to be executed properly to succeed.  We focus on the ALFRED environment and its defined tasks.

\noindent \textbf{Training a neural agent for VLN.} State-of-the-art models in language grounded navigation are neural agents trained using either Imitation Learning~\cite{fried2018speaker}, Reinforcement Learning~\cite{Li_2020_CVPR}, or a combination of both~\cite{tan-etal-2019-learning, rcm2019}. In addition, auxiliary tasks, such as progress estimation~\cite{ma2019regretful,ma2019selfmonitoring}, back-tracking~\cite{ke2019tactical}, speaker-driven route selection~\cite{fried2018speaker}, cross-modal matching~\cite{rcm2019,huang2019transferable}, back translation~\cite{tan-etal-2019-learning}, pretraining on subtasks~\cite{Zhu2020BabyWalkGF}, and text-based pretraining~\cite{cote18textworld,shridhar2020alfworld} are proposed to improve the performance and generalization of  neural agents in seen and unseen environments. Most of these approaches use recurrent neural networks and encode previous observations and actions as hidden states. Our work proposes to leverage transformers~\cite{attentionisall} which enables encoding the full episode of history for long-term navigation and interaction. Most relevant to our approach are VLN-BERT~\cite{majumdar2020improving} and Recurrent VLBERT~\cite{hong2020recurrent}, which also employ transformers for VLN. 
Unlike our approach, VLN-BERT~\cite{majumdar2020improving} trains a transformer to measure the compatibility of an instruction and a set of already generated trajectories. Concurrently, Recurrent VLBERT~\cite{hong2020recurrent} uses an explicit recurrent state and a pretrained VLBERT to process one observation for each timestep, which might have difficulty solving long-horizon tasks~\cite{NIPS2015_277281aa} such as ALFRED.
In contrast, we do not introduce any recurrency and process all the history of observations at once.

\noindent \textbf{Multimodal Transformers.}  Transformers~\cite{attentionisall} have brought success to a wide range of classification and generation tasks, from language~\cite{attentionisall,devlin2018bert,brown2020language} to images~\cite{dosovitskiy2020image,carion2020end} and videos~\cite{girdhar2019video,wang2018non}. In~\cite{pmlr-v119-parisotto20a}, the authors show that training transformers for long time horizon planning with RL is challenging and propose a solution. The convergence of the transformer architecture for different problem domains also leads to multimodal transformers, where a unified transformer model is tasked to solve problems that require multimodal information, such as visual question answering~\cite{lu2019vilbert}, video captioning and temporal prediction~\cite{Sun_2019_ICCV}, or retrieval~\cite{gabeur2020multi}. Our \model can be considered a multimodal transformer, where the inputs are language (instructions), vision (images), and actions.

\noindent \textbf{Semantic parsing of human instructions.} Semantic parsing focuses on converting natural language into logic forms that can be interpreted by machines. It has applications in question answering~\cite{zelle1996learning,zettlemoyer2007online,berant2013semantic} and can be learned either with paired supervision~\cite{zettlemoyer2012learning,yu2018spider,berant2013semantic} or weak supervision~\cite{artzi2013weakly,Patel-RSS-20}. For instruction following, semantic parsing has been applied to map natural language into lambda calculus expressions~\cite{artzi2013weakly} or linear temporal logic~\cite{Patel-RSS-20}. We show that rather than directly using the semantic parsing outputs, it is more beneficial to transfer its pretrained language encoder to the downstream VLN task.

\section{Method}
\label{sec:3}

We first define the vision-and-language navigation task in Section~\ref{sec:3.1} and describe the Episodic Transformer (E.T.) model in Section~\ref{sec:3.2}.
We then introduce the synthetic language and explain how we leverage it for pretraining and joint training in Section~\ref{sec:3.3}.

\subsection{VLN background}
\label{sec:3.1}
The vision-and-language navigation task requires an agent to navigate in an environment and to reach a goal specified by a natural language instruction.
Each demonstration is a tuple $(x_{1:L}, v_{1:T}, a_{1:T})$ of a natural language instruction, expert visual observations, and expert actions.
The instruction $x_{1:L}$ is a sequence of $L$ word tokens $x_i \in 
\mathbb{R}$.
The visual observations $v_{1:T}$ is a sequence of $T$ camera images $v_t \in \mathbb{R}^{W \times H \times 3}$ where $T$ is the demonstration length and $W \times H$ is the image size.
The expert actions $a_{1:T}$ is a sequence of $T$ action type labels $a_t \in \{1, \ldots, A\}$ used by the expert and $A$ is the number of action types.

The goal is to learn an agent function $f$ that approximates the expert policy. In the case of a recurrent architecture, the agent predicts the next action $\hat{a}_t$ given a language instruction $x_{1:L}$, a visual observation $v_t$, the previously taken action $\hat{a}_{t-1}$, and uses its hidden state $h_{t-1}$ to keep track of the history:
\begin{equation}
    \hat{a}_t, h_t = f(x_{1:L}, v_t, \hat{a}_{t-1}, h_{t-1}).
\end{equation}

For an agent with full episode observability, all previous visual observations $v_{1:t}$ and all previous actions $\hat{a}_{1:t-1}$ are provided to the agent directly and no hidden state is required:
\begin{equation}
    \hat{a}_t = f(x_{1:L}, v_{1:t}, \hat{a}_{1:t-1}).
    \label{eq:agent}
\end{equation}

\begin{figure*}[t!]
\centering
\includegraphics[width=\linewidth]{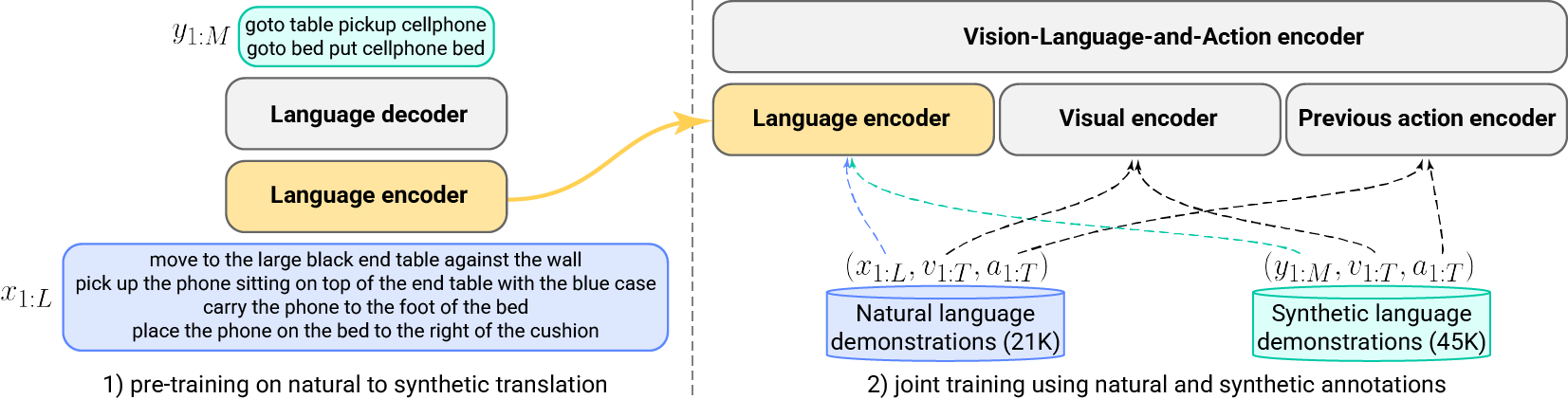}
\caption{Training with natural and synthetic language. 
Left: We pretrain the language encoder of the model to translate natural language instructions to synthetic language.
Due to a more task-oriented synthetic representation, the language encoder learns better representations.
We use the language encoder weights to initialize the language encoder of the agent (shown in yellow).
Right: We jointly use demonstrations annotated with natural language and demonstrations annotated with synthetic language to train the agent. Due to the larger size of the synthetic language dataset, the resulting agent has better performance even when evaluated on natural language annotations.
}
\label{fig:training}
\vspace{-0.5cm}
\end{figure*}

\subsection{Episodic Transformer model}
\label{sec:3.2}

Our Episodic Transformer (E.T.) model shown in Figure~\ref{fig:archi} relies on attention-based multi-layer transformer encoders~\cite{attentionisall}. It has no hidden state and observes the full history of visual observations and previous actions.
To inject information about the order of words, frames, and action sequences, we apply the sinusoidal encoding to transformer inputs. We refer to this encoding as positional encoding for language tokens and temporal encoding for expert observations and actions.

Our E.T. architecture consists of four encoders: language encoder, visual encoder, action encoder, and multimodal encoder.
The language encoder shown in the bottom-left part of Figure~\ref{fig:archi} gets instruction tokens $x_{1:L}$ as input. It consists of a look-up table and a multi-layer transformer encoder and outputs a sequence of contextualized language embeddings $h^x_{1:L}$.
The visual encoder shown in the bottom-center part of Figure~\ref{fig:archi} is a ResNet-50 backbone~\cite{resnet2016} followed by 2 convolutional and 1 fully-connected layers. The visual encoder projects a visual observation $v_t$ into its embedding $h^v_t$.
All the episode visual observations $v_{1:T}$ are projected independently using the same encoder.
The action encoder is a look-up table shown in the bottom-right part of Figure~\ref{fig:archi} which maps action types $a_{1:T}$ into action embeddings $h^a_{1:T}$.

The multimodal encoder is a multi-layer transformer encoder shown in the middle of Figure~\ref{fig:archi}.
Given the concatenated embeddings from modality-specific encoders $(h^x_{1:L}, h^v_{1:T}, h^a_{1:T})$, the multimodal encoder returns output embeddings $(z^x_{1:L}, z^v_{1:T}, z^a_{1:T})$.
The multimodal encoder employs causal attention~\cite{attentionisall} to prevent visual and action embeddings from attending to subsequent timesteps. We take the output embeddings $z^v_{1:T}$ and add a single fully-connected layer to predict agent actions $\hat{a}_{1:T}$.

During E.T. training, we take advantage of the sequential nature of the transformer architecture. We input a language instruction $x_{1:L}$ as well as all visual observations $v_{1:T}$ and all actions $a_{1:T}$ of an expert demonstration to the model. The E.T. model predicts all actions $\hat{a}_{1:T}$ at once as shown at the top of Figure~\ref{fig:archi}. We compute and minimize the cross-entropy loss between predicted actions $\hat{a}_{1:T}$ and expert actions $a_{1:T}$.
During testing at timestep $t$, we input visual observations $v_{1:t}$ up to a current timestep and previous actions $\hat{a}_{1:t-1}$ taken by the agent.
We select the action predicted for the last timestep $\hat{a}_t, \hat{c}_t$ and apply it to the environment which generates the next visual observation $v_{t+1}$. In Figure~\ref{fig:archi} we show an example that corresponds to the $6^{\textrm{th}}$ timestep of an episode where the action \texttt{Left} will be taken next.

\subsection{Synthetic language}
\label{sec:3.3}

To improve understanding of human instructions that present a wide range of variability,
we propose to pretrain the agent language encoder with a translation into a synthetic language, see Figure~\ref{fig:training} (left). We also generate additional demonstrations, annotate them with synthetic language and jointly train the agent using both synthetic and natural language demonstrations, see Figure~\ref{fig:training} (right).

An example of the synthetic language and a corresponding natural language instruction is shown in Figure~\ref{fig:training} (left).
The synthetic annotation is generated for each expert demonstration using the expert path planner arguments. In ALFRED, each expert path is defined with Planning Domain Definition Language (PDDL)~\cite{pddl1998} which consists of several subgoal actions. Each subgoal action has a type and a target class, e.g. \texttt{Put Apple Table} or \texttt{Goto Bed} which we use as a synthetic annotation for this subgoal action. Note that such annotation only defines a class but not an instance of the target.
We annotate each expert demonstration with subgoal action annotations concatenated in chronological order to produce a synthetic annotation $y_{1:M}$.

We use synthetic language to pretrain the language encoder of the agent on a sequence-to-sequence (seq2seq) translation task.
The translation dataset consists of corresponding pairs $(x_{1:L}, y_{1:M})$ of natural and synthetic instructions.
The translation model consists of a language encoder and a language decoder as shown in Figure~\ref{fig:training} (left). The language encoder is identical to the agent language encoder described in Section~\ref{sec:3.2}. The language decoder is a multi-layer transformer decoder with positional encoding and the same hyperparameters as the encoder.
Given a natural language annotation $x_{1:L}$, we use the language encoder to produce embeddings $h_{1:L}$.
The embeddings are passed to the language decoder which predicts $N$ translation tokens $\hat{y}_i$.
We train the model by minimizing the cross-entropy loss between predictions $\hat{y}_{1:N}$ and synthetic annotations $y_{1:M}$.
Once the training converges, we use the weights of the translator language encoder to initialize the language encoder of the agent.

We also explore joint training by generating an additional dataset of expert demonstrations annotated with synthetic language. We use the AI2-THOR simulator~\cite{ai2thor} and scripts provided by Shridhar \etal~\cite{ALFRED20}.
Apart from the annotations, the synthetic dataset differs from the original one in terms of objects configurations and agent initial positions.
We train the agent to predict actions using both natural and synthetic language datasets as shown on the right in Figure~\ref{fig:training}. We use the same language, vision, and action encoders for both datasets but two different look-up tables for natural and synthetic language tokens which we found to work the best experimentally.
For both datasets, we sample batches of the same size, compute the two losses and do a single gradient descent step. 
After a fixed number of training epochs, we evaluate the agent on natural and synthetic language separately using the same set of validation tasks.

\section{Results}
\label{sec:4}

In this section, we ablate different components of E.T. and compare E.T. with state-of-the-art methods. First, we describe the experimental setup and the dataset in Section~\ref{sec:4.1}. Next, we compare our method to a recurrent baseline and highlight the importance of full episode observability in Section~\ref{sec:4.2}.
We then study the impact of joint training and pretraining with synthetic instructions in Section~\ref{sec:4.3} and compare with previous state-of-the-art methods on the ALFRED dataset in Section~\ref{sec:4.4}.

\subsection{Experimental setup}
\label{sec:4.1}

\noindent \textbf{Dataset.} The ALFRED dataset~\cite{ALFRED20} consists of demonstrations of an agent performing household tasks following goals defined with natural language. The tasks are compositional with nonreversible state changes.
The dataset includes $8,055$ expert trajectories $(v_{1:T}, a_{1:T})$ annotated with $25,743$ natural language instructions $x_{1:L}$. It is split into $21,023$ train, $1,641$ validation, and $3,062$ test annotations. The validation and test folds are divided into \textit{seen} splits which contain environments from the train fold and \textit{unseen} splits which contain new environments. To leverage synthetic instructions to pretrain a language encoder, we pair every annotated instruction $x_{1:L}$ with its corresponding synthetic instruction $y_{1:M}$ in the train fold. For joint training, we generate $44,996$ demonstrations $(y_{1:M}, v_{1:T}, a_{1:T})$ from the train environments annotated automatically with synthetic instructions. For ablation studies in Section~\ref{sec:4.2} and Section~\ref{sec:4.3}, we use the validation folds only.
For comparison with state-of-the-art in Section~\ref{sec:4.4}, we report results on both validation and test folds.

\noindent \textbf{Baselines.}
In Section~\ref{sec:4.2}, we compare our model to a model based on a bi-directional LSTM~\cite{ALFRED20}.
We use the same hyperparameters as Shridhar \etal~\cite{ALFRED20} and set the language encoder hidden size to $100$, the action decoder hidden size to $512$, the visual embeddings size to $2500$, and use $0.3$ dropout for the decoder hidden state. We experimentally find the Adam optimizer with no weight decay and a weight coefficient $0.1$ for the target class cross-entropy loss to work best.
The LSTM model uses the same visual encoder as the E.T. model.
In Section~\ref{sec:4.4}, we also compare our model to MOCA~\cite{singh2020moca} and the model of Nguyen \etal~\cite{ngyuen_eval_winner}.

\noindent \textbf{Evaluation metrics.} For our ablation studies in Sections \ref{sec:4.2} and \ref{sec:4.3}, we report agent success rates. To understand the performance difference with recurrent-based architectures in Section~\ref{sec:4.2}, we also report success rates on individual subgoals. This metric corresponds to the proportion of subgoal tasks completed after following an expert demonstration until the beginning of the subgoal and conditioned on the entire language instruction. We note that the average task length is $50$ timesteps while the average length of a subgoal is $7$ timesteps.

\noindent \textbf{Implementation details.}
Among the $13$ possible action types, $7$ actions involve interacting with a target object in the environment. The target object of an action $a_t$ is chosen with a binary mask $m_t \in \{0, 1\}^{W \times H}$ that specifies the pixels of visual observation $v_t$ that belong to the target object. There are $119$ object classes in total. The pixel masks $m_t$ are provided along with expert demonstrations during training. We follow Singh \etal~\cite{singh2020moca} and ask our agent to predict the target object class $c_t$, which is then used to retrieve the corresponding pixel mask $\hat{m}_t$ generated by a pretrained instance segmentation model. The segmentation model takes $v_t$ as input and outputs $(\hat{c}_t, \hat{m}_t)$.

The agent observations are resized to $224 \times 224$. The mask generator receives images of size $300 \times 300$ following Singh \etal~\cite{singh2020moca}.
Both the visual encoder and the mask generator are pretrained on a dataset of $325$K frames of expert demonstrations from the train fold and corresponding class segmentation masks.
We use ResNet-50 Faster R-CNN~\cite{fasterrcnn2015} for the visual encoder pretraining and ResNet-50 Mask R-CNN~\cite{maskrcnn2017} for the mask generator.
We do not update the mask generator and the visual encoder ResNet backbone during the agent training.
In the visual encoder, ResNet features are average-pooled $4$ times to reduce their size and $0.3$ dropout is applied. Resulting feature maps of $512 \times 7 \times 7$ are fed into 2 convolutional layers with $256$ and $64$ filters of size $1$ by $1$ and mapped into an embedding of the size $768$ with a fully connected layer.
Both transformer encoders of E.T. have $2$ blocks, $12$ self-attention heads, and the hidden size of $768$.
We use $0.1$ dropout inside transformer encoders.

We use the AdamW optimizer~\cite{loshchilov2018decoupled} with $0.33$ weight decay and train the model for $20$ epochs. Every epoch includes $3,750$ batches of $8$ demonstrations each.
For joint training, each batch consists of $4$ demonstrations with human instructions and $4$ demonstrations with synthetic instructions.
For all experiments, we use a learning rate of $10^{-4}$ during the first $10$ epochs and $10^{-5}$ during the last $10$ epochs.
Following Shridhar \etal~\cite{ALFRED20}, we use auxiliary losses for overall and subgoal progress~\cite{ma2019selfmonitoring} which we sum to the model cross-entropy loss with weights $0.1$.
All the hyperparameter choices were made using a moderate size grid search.
Once the training is finished, we evaluate every $2$-nd epoch on the validation folds.
Following Singh \etal~\cite{singh2020moca}, we use Instance Association in Time and Obstruction Detection modules during evaluation.

\subsection{Model analysis}
\label{sec:4.2}
\begin{table}
\centering
\begin{tabular}{lacac}
\toprule
\multirow{2}{*}{Model}
& \multicolumn{2}{c}{Task} & \multicolumn{2}{c}{Sub-goal} \\
& Seen & Unseen & Seen & Unseen \\
\midrule
LSTM & $23.2$ & $2.4$ & $75.5$ & $58.7$ \\
LSTM + E.T. enc. & $27.8$ & $\B{3.3}$ & $76.6$ & $59.5$ \\
E.T. & \B{33.8} & $3.2$ & $\B{77.3}$ & $\B{59.6}$ \\
\bottomrule
\end{tabular}
\caption{Comparison of E.T. and LSTM architectures: (1) an LSTM-based model~\cite{ALFRED20}, (2) an LSTM-based model trained with the transformer language encoder of the E.T. model, (3) E.T., our transformer-based model. All models are trained using the natural language dataset only and evaluated on validation folds. The two parts of the table show the success rate for tasks (average length $50$) and sub-goals (average length $7$).
While the sub-goal success rates of all models are relatively close, E.T.
outperforms both recurrent agents on full tasks which highlights the importance
of the full episode observability.}
\label{tab:4.2.archi}
\vspace{-0.1cm}
\end{table}

\begin{table}
\centering
\begin{tabular}{lacac}
\toprule
    \multirow{2}{*}{Visible}
    & \multicolumn{2}{c}{Frames} & \multicolumn{2}{c}{Actions} \\
    & Seen & Unseen & Seen & Unseen \\
\midrule
None & $0.5$ & $0.2$ & $23.7$ & $1.7$ \\
1 last & $28.9$ & $2.2$ & $\B{33.8}$ & $\B{3.2}$ \\
4 last & $31.5$ & $2.0$ & $32.0$ & $2.4$ \\
16 last & $33.5$ & $2.9$ & $31.1$ & $2.8$ \\
All & $\B{33.8}$ & $\B{3.2}$ & $27.1$ & $2.2$ \\
\bottomrule
\end{tabular}
\caption{Ablation on accessible history length of E.T., in terms of visual frames (left two columns) and actions (right two columns).}
\label{tab:4.2.memory}
\vspace{-0.4cm}
\end{table}

\noindent \textbf{Comparison with recurrent models.}
To validate the gain due to the episodic memory, we compare the E.T. architecture with a model based on a recurrent LSTM architecture.
We train both models using the dataset with natural language annotations only.
As shown in Table~\ref{tab:4.2.archi}, the recurrent model succeeds in $23.2\%$ of tasks in seen environments and in $2.4\%$ of tasks in unseen environments. E.T. succeeds in $33.8\%$ and $3.2\%$ of tasks respectively which is a relative improvement of $45.6\%$ and $33.3\%$ compared to the LSTM-based agent. However, the success rate computed for individual subgoals shows only $2.3\%$ and $1.5\%$ of relative improvement of E.T. over the recurrent agent in seen and unseen environments respectively. We note that a task consists on average of $6.5$ subgoals which makes the long-term memory much more important for solving full tasks.

To understand the performance difference, we train an LSTM-based model with the E.T. language encoder.
Given that both LSTM and E.T. agents receive the same visual features processed by the frozen ResNet-50 backbone and have the same language encoder architecture, the principal difference between the two models is the processing of previous observations. While the E.T. agent observes all previous frames using the attention mechanism, the LSTM-based model relies on its recurrent state and explicitly observes only the last visual frame. The recurrent model performance shown in the $2$-nd row of Table~\ref{tab:4.2.archi} is similar to the E.T. performance in unseen environments but is $17.7\%$ less successful than E.T. in seen environments. This comparison highlights the importance of the attention mechanism and full episode observability.
We note that E.T. needs only one forward pass for a gradient descent update on a full episode.
In contrast, the LSTM models need to do a separate forward pass for each episode timestep which significantly increases their training time with respect to E.T. models.
We further compare how E.T. and LSTM models scale with additional demonstrations in Section~\ref{sec:4.3}.

\noindent \textbf{Accessible history length.}
We train E.T. using different lengths of the episode history observed by the agent in terms of visual frames and previous actions and show the results in Table~\ref{tab:4.2.memory}. The first two columns of Table~\ref{tab:4.2.memory} compare different lengths of visual observations history from no past frames to the entire episode. The results indicate that having access to all  visual observations is important for the model performance. We note that the performance of the model with $16$ input frames is close to the performance of the full episode memory agent, which can be explained by the average task length of $50$ timesteps.

The last two columns of Table~\ref{tab:4.2.memory} show that the agent does not benefit from accessing more than one past action. This behavior can be explained by the ``causal misidentification'' phenomenon: access to more information can yield worse performance~\cite{de2019causal}. It can also be explained by poor generalizability due to the overfitting of the model to expert demonstrations. We also note that the model observing no previous actions is $29.8\%$ and $46.8\%$ relatively less successful in seen and unseen environments than the agent observing the last action. We, therefore, fix the memory size to be unlimited for visual observations and to be $1$ timestep for the previous actions.

\noindent \textbf{Model capacity.}
Transformer-based models are known to be expressive but prone to overfitting. We study how the model capacity impacts the performance while training on the original ALFRED dataset.
We change the number of transformer blocks in the language encoder and the multimodal encoder and report results in Table~\ref{tab:4.2.layers}. The results indicate that the model with a single transformer block is not expressive enough and the models with $3$ and $4$ blocks overfit to the train data. The model with $2$ blocks represents a trade-off between under- and overfitting and we, therefore, keep this value for all the experiments.

\noindent \textbf{Attention visualization.} We visualize text and visual attention heatmaps in Appendices~\textcolor{red}{A.4} and~\textcolor{red}{A.5} of~\cite{et2021}.

\subsection{Training with synthetic annotations}
\label{sec:4.3}
\begin{table}
\centering
\begin{tabular}{cac}
\toprule
    $\#$ Blocks & Seen & Unseen \\
\midrule
1 & $25.0$ & $1.6$ \\
2 & $\B{33.8}$ & $\B{3.2}$ \\
3 & $28.6$ & $2.2$ \\
4 & $19.8$ & $1.1$ \\
\bottomrule
\end{tabular}
\caption{Ablation of E.T. model capacity. We compare E.T. models with different number of transformer blocks in language and multimodal encoders.}
\label{tab:4.2.layers}
\vspace{-0.15cm}
\end{table}

\begin{table}
\setlength\tabcolsep{2.7pt}
\centering
\begin{tabular}{lacac}
\toprule
\multirow{2}{*}{Synthetic instr.} & \multicolumn{2}{c}{Test on synthetic} & \multicolumn{2}{c}{Test on human} \\
& Seen & Unseen & Seen & Unseen \\
\midrule
Expert frames & $\B{54.0}$ & $\B{6.1}$ & $28.5$ & $3.4$ \\
Speaker text & $36.3$ & $3.1$ & $37.4$ & $3.9$ \\
Subgoal actions & $47.2$ & $5.9$ & $\B{38.5}$ & $\B{5.4}$ \\
\cmidrule(lr){1-5}
No synthetic & \multicolumn{1}{b}{-} & - & $33.8$ & $3.2$ \\
\bottomrule
\end{tabular}
\caption{Comparison of different synthetic instructions used for \textbf{joint training}.
We jointly train E.T. using demonstrations with human annotations and demonstrations with different types of synthetic instructions.
In the first two columns, we evaluate the resulting models using the same type of synthetic annotations that is used during training. 
In the last two columns, the models are evaluated on human annotated instructions.
}
\label{tab:4.3.annotations}
\vspace{-0.15cm}
\end{table}

\begin{table}
\centering
\begin{tabular}{@{}lacacacac@{}}
\toprule
    \multirow{2}{*}{Train data}
    & \multicolumn{2}{c}{{LSTM}} & \multicolumn{2}{c}{{E.T.}} \\
    & \multicolumn{1}{a}{Seen} & \multicolumn{1}{c}{Unseen}  & \multicolumn{1}{a}{Seen} & \multicolumn{1}{c}{Unseen} \\
                 
\midrule
Human annotations & $23.2$ & $2.4$ & $33.8$ & $3.2$ \\
Human $+$ synthetic & $25.2$ & $2.9$ & $\B{38.5}$ & $\B{5.4}$ \\
\bottomrule
\end{tabular}
\caption{
Comparison of an LSTM-based model and E.T. \textbf{trained jointly with demonstrations annotated by subgoal actions}.
The results indicate that E.T. scales better with additional data than the LSTM-based agent.}
\label{tab:4.3.lstms}
\vspace{-0.4cm}
\end{table}

\noindent \textbf{Joint training.}
We train the E.T. model using the original dataset of $21,023$ expert demonstrations annotated with natural language and the additionally generated dataset of $44,996$ expert demonstrations with synthetic annotations.
We compare three types of synthetic annotations: (1)~direct use of visual embeddings from the expert demonstration frames, no language instruction is generated. A similar approach can be found in Lynch and Sermanet~\cite{lynch2020language}; (2)~train a model to generate instructions, \eg with a speaker model~\cite{fried2018speaker}, where the inputs are visual embeddings from the expert demonstration frames, and the targets are human-annotated instructions; and (3)~subgoal actions and objects annotations described in Section~\ref{sec:3.3}. For (1), we experimentally find using all expert frames from a demonstration works significantly better than a subset of frames. The visual embeddings used in (1) and (2) are extracted from a pretrained frozen ResNet-50 described in Section~\ref{sec:4.1}. To generate speaker annotations, we use a transformer-based seq2seq model (Section~\ref{sec:3.3}) with the difference that the inputs are visual embeddings instead of text.

We report success rates of models trained jointly and evaluated independently on synthetic and human-annotated instructions in Table~\ref{tab:4.3.annotations}. The results are reported on the validation folds.
The model trained on expert frames achieves the highest performance when evaluated on synthetic instructions.
However, when evaluated on human instructions, this model has $15.6\%$ relatively lower success rate in seen environments than the baseline without joint training. This indicates that the agent trained to take expert frames as instructions does not generalize well to human instructions.
Using speaker translation annotations improves over the no joint training baseline by $10.6\%$ and $21.8\%$ in seen and unseen environments respectively. Furthermore, our proposed subgoal annotations bring an even larger relative improvement of $13.9\%$ and $68.7\%$ in seen and unseen environments which highlights the benefits of joint training with synthetic instructions in the form of subgoal actions.

Finally, we study if the recurrent baseline also benefits from joint training with synthetic data.
Table~\ref{tab:4.3.lstms} shows that the relative gains of joint training are $2.3$ and $4.4$ times higher for E.T. than for the LSTM-based agent in seen and unseen environments respectively.
These numbers clearly show that E.T. benefits more from additional data and confirms the advantage of our model over LSTM-based agents.

\noindent \textbf{Language encoder pretraining.} Another application of synthetic instructions is to use them as an intermediate representation that decouples the visual appearance of an environment from the variations of human-annotated instructions. For this purpose, we pretrain the E.T. language encoder with the synthetic instructions. In particular, we pretrain a seq2seq model to map human instructions into synthetic instructions as described in Section~\ref{sec:3.3}, and study whether it is more beneficial to transfer explicitly the ``translated'' text or implicitly as representations encoded by the model weights. Our pretraining is done on the original train fold with no additionally generated trajectories. The seq2seq translation performance is very competitive, reaching $97.1\%$ in terms of F1 score. To transfer explicitly the translated (synthetic) instructions, we first train an E.T. agent to follow synthetic instructions on the training fold and then evaluate the agent on following human instructions by translating these instructions into synthetic ones with our pretrained seq2seq model.

Table~\ref{tab:4.3.pretrain} compares these two pretraining strategies. We can see that both strategies outperform the no pretraining baseline (first row) significantly and that transferring the encoder works better than explicit translation.
For completeness, we also report results with BERT pretraining~\cite{devlin2018bert} (second row). The BERT model is pretrained on generic text data (\eg Wikipedia). We use the BERT base model whose weights are released by the authors. We extract its output contextualized word embeddings and use them as the input word embeddings to the language encoder.
To our surprise, when compared with the no pretraining baseline, the BERT pretraining decreases the performance in seen environments by $4.4\%$ and brings a marginal improvement of $6.2\%$ relative in unseen environments. We conjecture that domain-specific language pretraining is important for the ALFRED benchmark. Overall, these experiments show another advantage of the proposed synthetic annotations and highlight the importance of intermediate language representations to better train instruction-following agents.

We finally combine the language encoder pretraining and the joint training objectives and present the results in Table~\ref{tab:4.3.joint_and_pretrain}. We observe that these two strategies are complementary to each other: the overall relative improvements of incorporating synthetic data over the baseline E.T. model are $37.8\%$ and $228.1\%$ in seen and unseen environments, respectively. We conclude that synthetic data is especially important for generalization to unseen environments.
A complete breakdown of performance improvements can be found in Appendix~\textcolor{red}{A.2} of~\cite{et2021}.

\begin{table}
\centering
\begin{tabular}{ccac}
\toprule
Objective & Transfer  & Seen & Unseen \\
\midrule
None & - &  $33.8$ & $3.2$ \\
BERT & Text embedding & $32.3$ & $3.4$ \\
Seq2seq & Translated text & $35.2$ & $3.6$ \\
Seq2seq & Text encoder & $\B{37.6}$ & $\B{3.8}$ \\
\bottomrule
\end{tabular}
\caption{
Comparison of models \textbf{with different language encoder pretraining} strategies. We pretrain a seq2seq model to map human instructions into synthetic instructions and transfer either its output text (third row) or its learned weights (fourth row). For completeness, we also compare with no pretraining (first row) and BERT pretraining (second row).}
\label{tab:4.3.pretrain}
\vspace{-0.4cm}
\end{table}

\begin{table}[t!]
\centering
\begin{tabular}{@{}ccac@{}}
\toprule
  Pretraining & Joint training & Seen & Unseen \\
\midrule
& & $33.8$ & $3.2$ \\
\cmidrule(lr){1-4}
\cmark & & $37.6$ & $3.8$ \\
& \cmark & $38.5$ & $5.4$ \\
\cmark & \cmark & $\B{46.6}$ & $\B{7.3}$ \\
\bottomrule
\end{tabular}
\caption{
Ablation study of \textbf{joint training and language encoder pretraining} with synthetic data. We present baseline results without leveraging synthetic data (first row), the independent performance of pretraining (second row) and joint training (third row), and their combined performance (fourth row).}
\label{tab:4.3.joint_and_pretrain}
\vspace{-0.2cm}
\end{table}

\begin{table}[t!]
    \centering
    \begin{tabular}{@{}lacac@{}}
        \toprule
            \multirow{2}{*}{Model}
            & \mcc{2}{\textbf{Validation}} & \mcc{2}{\textbf{Test}} \\
            & Seen & Unseen & Seen & Unseen \\
        \midrule
            {Shridhar \etal~\cite{ALFRED20}} & $3.70$ & $0.00$ & $3.98$ & $0.39$ \\[1pt]
            {Nguyen \etal~\cite{ngyuen_eval_winner}} & \multicolumn{1}{b}{N/A} & \multicolumn{1}{c}{N/A} & $12.39$ & $4.45$ \\[1pt]
            {Singh \etal~\cite{singh2020moca}} & $19.15$ & $3.78$ & $22.05$ & $5.30$ \\[1pt]
            {E.T.} & $33.78$ & $3.17$ & $28.77$ & $5.04$ \\[1pt]
            {E.T. (pretr.)} & $37.63$ & $3.76$ & $33.46$ & $5.56	$\\[1pt]
            {E.T. (pretr. \& joint tr.)} & $\B{46.59}$ & $\B{7.32}$ & $\B{38.42}$ & $\B{8.57}$ \\[1pt]
        \midrule
            {Human performance}     & \multicolumn{1}{b}{-} & \multicolumn{1}{c}{-} & \multicolumn{1}{b}{-} & $91.00$ \\
        \bottomrule
        \end{tabular}
    \caption{
    Comparison with the models submitted to the public leaderboard on validation and test folds.
    The highest value per fold is shown in \B{blue}.
    `N/A' denotes that the scores are not reported on the leaderboard or in an associated publication.
    Our method sets a new state-of-the-art on all metrics.
    }
    \label{tab:4.4.sota}
\vspace{-0.4cm}
\end{table}

\subsection{Comparison with state-of-the-art}
\label{sec:4.4}

We compare the E.T. agent with models with associated tech reports on the public leaderboard\footnote{\url{https://leaderboard.allenai.org/alfred}, the results were submitted on February 22, 2021.}. 
The results on validation and test folds are shown in Table~\ref{tab:4.4.sota}.
The complete table with \textit{solved goal conditions} and
\textit{path-length-weighted scores}~\cite{anderson2019evaluation} is given in
Appendix~\textcolor{red}{A.1} of~\cite{et2021}~.
The E.T. model trained without synthetic data pretraining and joint training
sets a new state-of-the-art on seen environments (row 4). By leveraging
synthetic instructions for pretraining, our method outperforms the
previous methods~\cite{ALFRED20, ngyuen_eval_winner, singh2020moca} and sets a
new state-of-the-art on all metrics (row 5). Given additional $45$K trajectories for joint training, the E.T. model further improves the results (row 6).

\section{Conclusion}
\label{sec:5.conclusion}
We propose E.T., a transformer-based architecture for vision-and-language navigation tasks. E.T. observes the full episode history of vision, language, and action inputs and encodes it with a multimodal transformer.
On the ALFRED benchmark, E.T. outperforms competitive recurrent baselines and achieves state-of-the-art performance on seen environments.
We also propose to use synthetic instructions for pretraining and joint training with human-annotated instructions.
Given the synthetic instructions, the performance is further improved in seen and especially, in unseen environments. In the future, we want to explore other forms of synthetic annotations and techniques to automatically construct them, for example  with object detectors.

~\\
\noindent\textbf{Acknowledgement:} We thank Peter Anderson, Ellie Pavlick, and Dylan Ebert for helpful feedback on the draft.

{\small
\bibliographystyle{ieee_fullname}
\bibliography{main}
}

\clearpage

\appendix
\renewcommand\thefigure{A\arabic{figure}}
\renewcommand\thetable{A\arabic{table}}
\setcounter{figure}{0}
\setcounter{table}{0}

\section{Appendix}
In this appendix we provide additional results and analysis, including the state-of-the-art comparison with all evaluation metrics, the complete breakdown of performance improvements, the impact of synthetic data size, and visualizations of transformer attention maps.

\subsection{Comparison with state of the art}
\label{app:sota}

We present a complete table comparing E.T. to the state-of-the-art methods from the public leaderboard.
In addition to the success rates on validation and test folds reported in the main paper (denoted as \textit{Full task}), we measure the amount of subgoal conditions completed for each task on average~\cite{ALFRED20} (denoted as \textit{Goal Cond.}).
We also compute path-length-weighted scores~\cite{anderson2019evaluation} for both metrics which weight the original metric value by the ratio of the agent path length and the expert path length~\cite{ALFRED20}.
Table~\ref{tab:6.sota_full} shows that the results on the additional metrics strongly correlate with the full task success rates reported in the main paper.

\begin{table}[!htbp]
    \centering
    \resizebox{1.00\linewidth}{!}{
        \begin{tabular}{laarr}
            \toprule
            \multirow{3}{*}{Model}
                & \mcp{4}{\textbf{Validation}} \\
                & \mcc{2}{\textit{Seen}}   & \mcc{2}{\textit{Unseen}}  \\
                & \multicolumn{1}{b}{Full task} & \multicolumn{1}{b}{Goal Cond.} 
                & \multicolumn{1}{c}{Full task} & \multicolumn{1}{c}{Goal Cond.} \\
            \cmidrule{1-5}
            {Shridhar \etal~\cite{ALFRED20}}        & $3.70$ ($2.10$)    & $10.00$  ($7.00$)    & $0.00$ ($0.00$)   & $6.90$ ($5.10$)  \\[1pt]
            {Nguyen \etal~\cite{ngyuen_eval_winner}}          & \multicolumn{1}{b}{N/A} & \multicolumn{1}{b}{N/A} & \multicolumn{1}{c}{N/A} & \multicolumn{1}{c}{N/A} \\[1pt]
            {Singh \etal~\cite{singh2020moca}}                                     & $19.15$ ($13.60$) & $28.50$  ($22.30$) & $3.78$ ($2.00$) & $13.40$ ($8.30$) \\[1pt]
            {E.T.} & $33.78$ ($24.90$) & $42.48$ ($33.10$) & $3.17$ ($1.34$) & $13.12$ ($7.41$) \\[1pt]
            {E.T. (pretr.)} & $37.63$ ($28.03$) & $47.59$ ($37.27$) & $3.76$ ($2.20$) & $14.65$ ($8.44$)\\[1pt]
            {E.T. (pretr. + \& joint tr.)} & $\B{46.59}$ ($\B{32.26}$) & $\B{52.82}$ ($\B{42.24}$) & $\B{7.32}$ ($\B{3.34}$) & $\B{20.87}$ ($\B{11.31}$) \\[1pt]
            \bottomrule
        \end{tabular}
    } \\
    \resizebox{1.00\linewidth}{!}{
        \begin{tabular}{laarr}
            \multirow{3}{*}{Model}
                &  \mcc{4}{\textbf{Test}} \\
                & \mcc{2}{\textit{Seen}}   & \mcc{2}{\textit{Unseen}}  \\
                & \multicolumn{1}{b}{Full task} & \multicolumn{1}{b}{Goal Cond.} 
                & \multicolumn{1}{c}{Full task} & \multicolumn{1}{c}{Goal Cond.} \\
            \cmidrule{1-5}
            {Shridhar \etal~\cite{ALFRED20}} & $3.98$ ($2.02$)   & $9.42$ ($6.27$)   & $0.39$ ($0.80$) & $7.03$ ($4.26$) \\[1pt]
            {Nguyen \etal~\cite{ngyuen_eval_winner}} & $12.39$ ($8.20$) & $20.68$ ($18.79$)   & $4.45$ ($2.24$) & $12.34$ ($9.44$) \\[1pt]
            {Singh \etal~\cite{singh2020moca}} & $22.05$ ($15.10$) & $28.29$ ($22.05$) & $5.30$ ($2.72$) & $14.28$ ($9.99$) \\[1pt]
            {E.T.} & $28.77$ ($19.77$) & $36.47$ ($28.00$) & $5.04$ ($1.94$) & $15.01$ ($8.73$) \\[1pt]
            {E.T. (pretr.)} & $33.46$ ($23.82$) & $41.08$ ($31.52$) & $5.56$ ($2.82$) & $15.44$ ($9.62)$\\[1pt]
            {E.T. (pretr. \& joint tr.)} & $\B{38.42}$ $(\B{27.78}$) & $\B{45.44}$ ($\B{34.93}$) & $\B{8.57}$ ($\B{4.10}$) & $\B{18.56}$ ($\B{11.46}$) \\[1pt]
            \cmidrule{1-5}
            {Human}     & \multicolumn{1}{b}{-} & \multicolumn{1}{b}{-} & $91.00$ ($85.80$) & $94.50$ ($87.60$) \\
            \bottomrule
        \end{tabular}
    }
    \captionsetup{type=table}
    \captionof{table}{Comparison with the models submitted to the public leaderboard on validation and test folds. We report success rates of the models on full tasks and subgoal conditions. We weight agent path lengths with expert path lengths and report path-length-weighted scores in parenthesis.
    The highest value per fold is shown in \B{blue}.
    `N/A' denotes that the scores are not reported on the leaderboard or in an associated publication.
    }
    \label{tab:6.sota_full}
\end{table}

\subsection{Complete performance analysis}
\label{app:complete_perf}
\begin{table}[!htb]
\centering
\begin{tabular}{@{}lacacacac@{}}
\toprule
    Components & \multicolumn{1}{a}{Seen} & \multicolumn{1}{c}{Unseen} \\
\midrule
LSTM baseline (Shridhar \etal~\cite{ALFRED20})& $4.8$ & $0.2$ \\
\cmidrule(lr){1-3}
+ ALFRED detection pretraining & $8.5$ & $0.4$ \\
+ Pretrained MaskRCNN~\cite{singh2020moca} & $23.2$ & $2.4$ \\
- LSTM; + Transformer (E.T.) & $33.8$ & $3.2$ \\
+ Synthetic language pretraining & $37.6$ & $3.8$ \\
+ Joint training with $45$K demonstrations & $\B{46.6}$ & $\B{7.3}$ \\
\bottomrule
\end{tabular}
\caption{
Complete breakdown of performance improvements. We report the performance of the model proposed by Shridhar \etal~\cite{ALFRED20} and sequentially add components that improve its success rate one by one.
The components include (1) visual features pretrained to detect objects in ALFRED, (2) a pretrained MaskRCNN to predict pixel masks, (3) the E.T. model, (4) language encoder pretraining on human to synthetic translation, (5) joint training with additional data.}
\label{tab:6.complete_perf}
\vspace{-0.4cm}
\end{table}

We present a complete breakdown of performance improvements with respect to the components added to the LSTM-based baseline model proposed by Shridhar \etal~\cite{ALFRED20}. First, we replace ImageNet visual features with features pretrained to detect objects in ALFRED as explained in Section~\ref{sec:4.1}. Next, we replace explicit pixel mask predictions with a pretrained MaskRCNN model proposed by Singh \etal~\cite{singh2020moca}. These two components combined bring a significant improvement over the original baseline performance~\cite{ALFRED20}. We then replace the LSTM model with the E.T. architecture, pretrain the language encoder of the agent to translate human language to synthetic representations, and jointly train the agent using additional $45$K demonstrations to achieve the state-of-the-art performance reported in Table~\ref{tab:6.sota_full}.

\subsection{Impact of synthetic demonstration size}
\label{app:synth_data}

We extend the results of Table~\ref{tab:4.3.lstms} and train the E.T. agent using different number of demonstrations annotated with synthetic instructions. The results are shown in Table~\ref{tab:6.synth_datas}. We can see that increasing the number of synthetic demonstrations in the joint training up to $22$K brings a significant improvement over the model trained on human annotations only. However, doubling the synthetic demonstrations up to $44$K has a very minor impact on the agent performance. We use $44$K synthetic data in the main paper.

\begin{table}[!htb]
\centering
\begin{tabular}{@{}lacacacac@{}}
\toprule
    Train data & \multicolumn{1}{a}{Seen} & \multicolumn{1}{c}{Unseen} \\
\midrule
21K human only & $33.8$ & $3.2$ \\
21K human \& 11K synth. & $35.5$ & $4.1$ \\
21K human \& 22K synth. & $38.3$ & $\B{5.5}$ \\
21K human \& 44K synth. & $\B{38.5}$ & $5.4$ \\
\bottomrule
\end{tabular}
\caption{
\textbf{Joint training} of the E.T. model using different number of demonstrations annotated with synthetic instructions. We report success rates on the validation folds.}
\label{tab:6.synth_datas}
\vspace{-0.4cm}
\end{table}

\subsection{Visualizing visual attention}
\label{app:attn_vis}
\begin{figure*}
\centering
\includegraphics[trim={0 1cm 0 0}, width=0.9\linewidth]{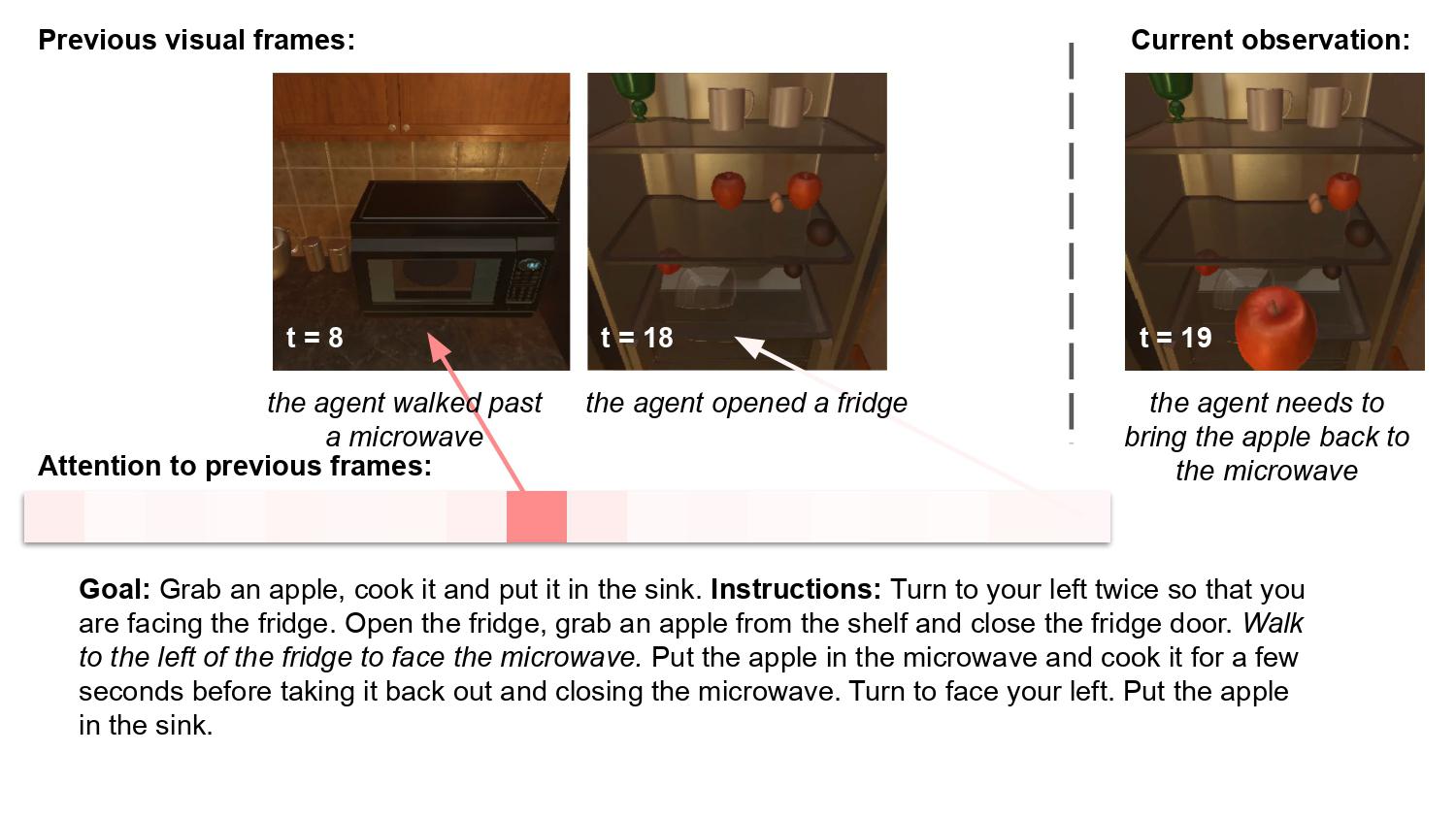}
\vspace{1cm}
\caption{A visualization of normalized attention heatmap to previous visual observations, from white (no attention) to red (high attention). In this example, a microwave is first observed at the $8^{\textrm{th}}$ timestep, and is highlighted by the visual attention at the $19^{\textrm{th}}$ timestep when the agent is asked to put the apple in the microwave.} 
\label{fig:attn_vis1}
\end{figure*}

\begin{figure*}
\centering
\includegraphics[trim={0 1cm 0 0}, width=0.9\linewidth]{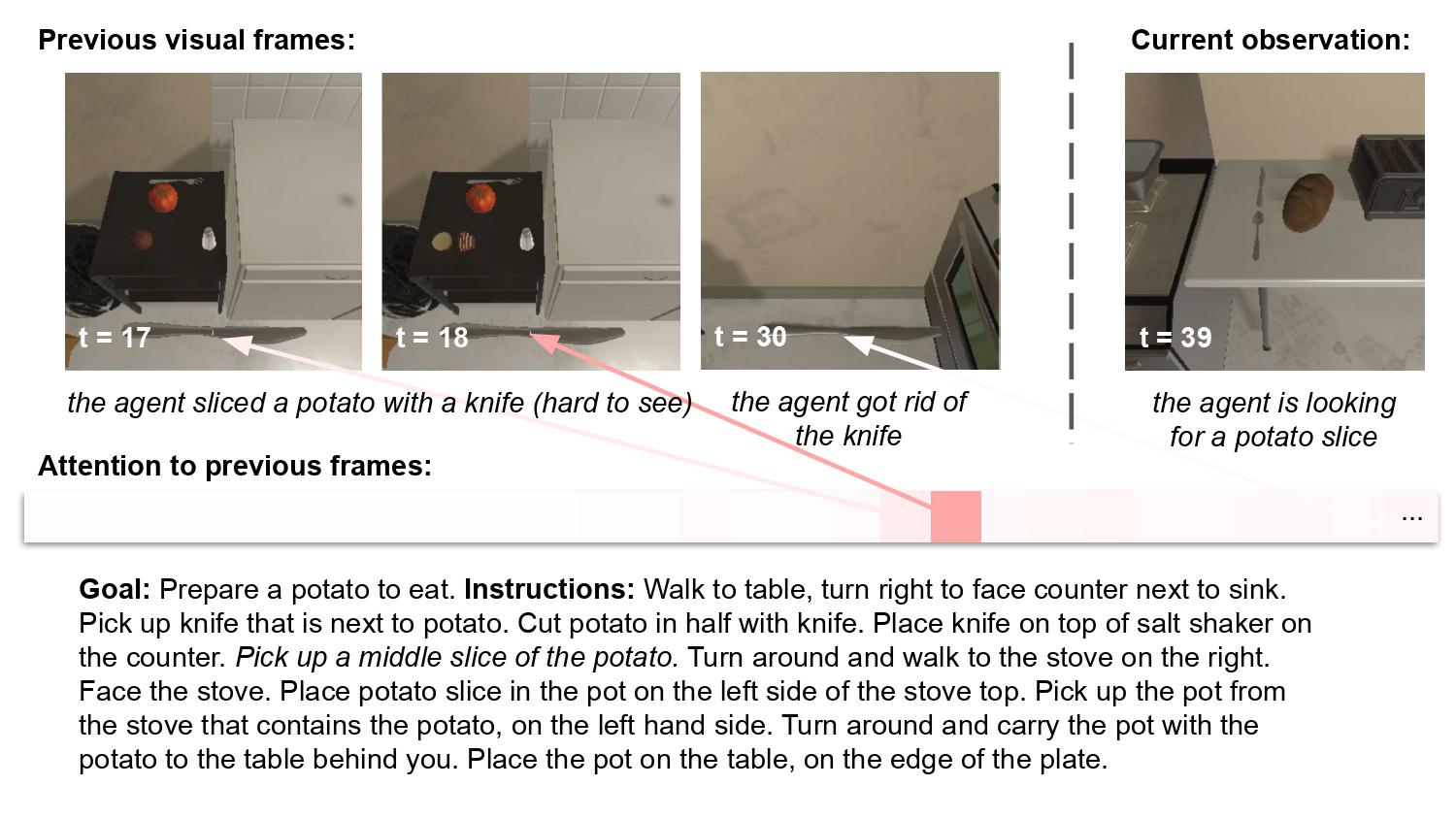}
\vspace{1cm}
\caption{A visualization of normalized attention heatmap to previous visual observations. In this example, the agent is asked to cut a potato (timesteps $17-18$) and to put a slice of it in a pot. At timestep $39$ when the agent is asked to retrieve the sliced potato, it attends to frames at timesteps $17-18$ to decide where to go.}
\label{fig:attn_vis2}
\end{figure*}

\begin{figure*}
\centering
\includegraphics[trim={0 1cm 0 0}, width=0.9\linewidth]{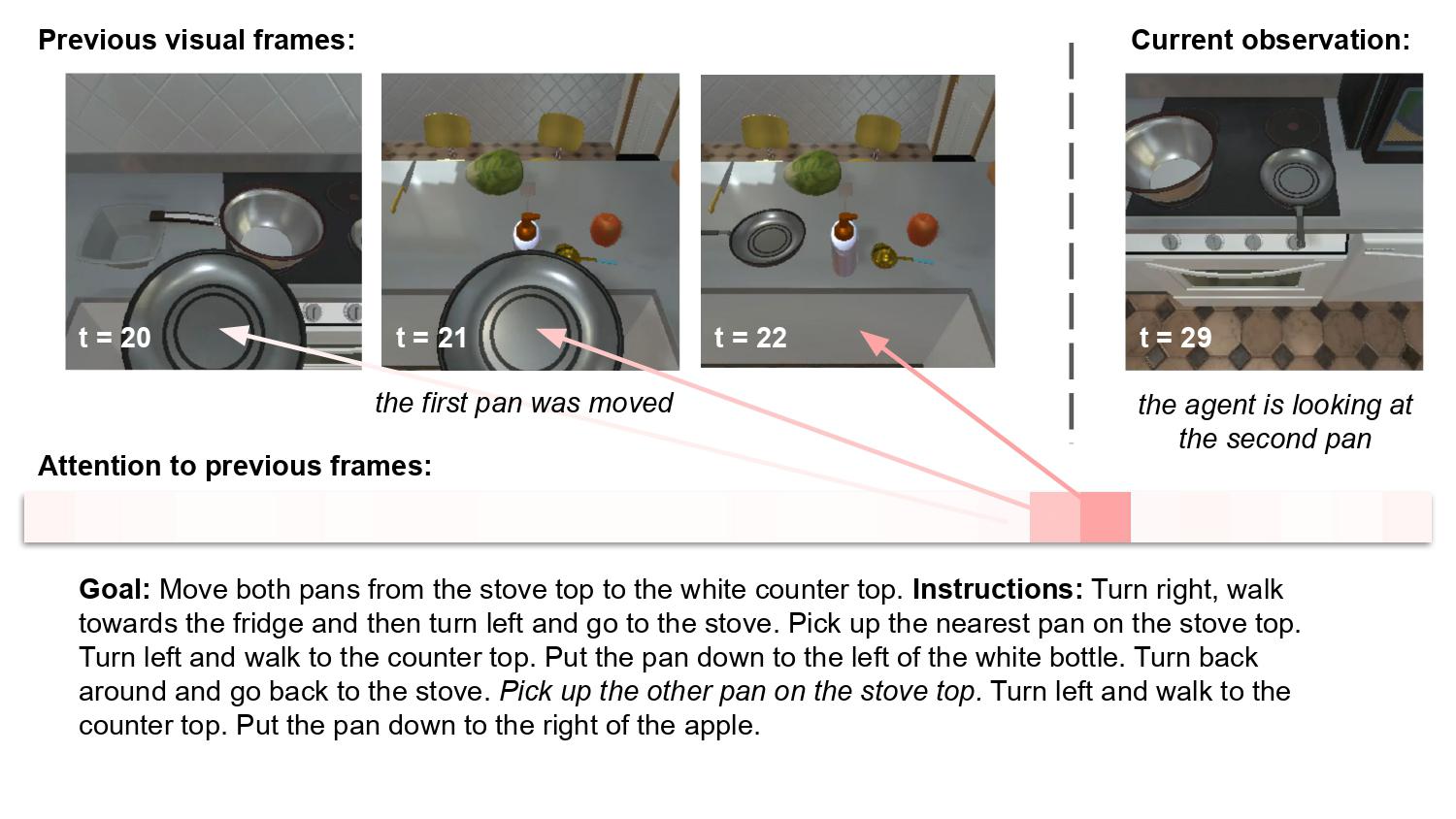}
\vspace{1cm}
\caption{A visualization of normalized attention heatmap to previous visual observations. In this example, the agent is asked to move two identical pans. It moves the first pan at timesteps $20-22$ and attends the frame at timestep $29$ when moving the second pan (see the two corresponding pink squares).}
\label{fig:attn_vis3}
\end{figure*}

\begin{figure*}
\centering
\includegraphics[trim={0 1cm 0 0}, width=0.9\linewidth]{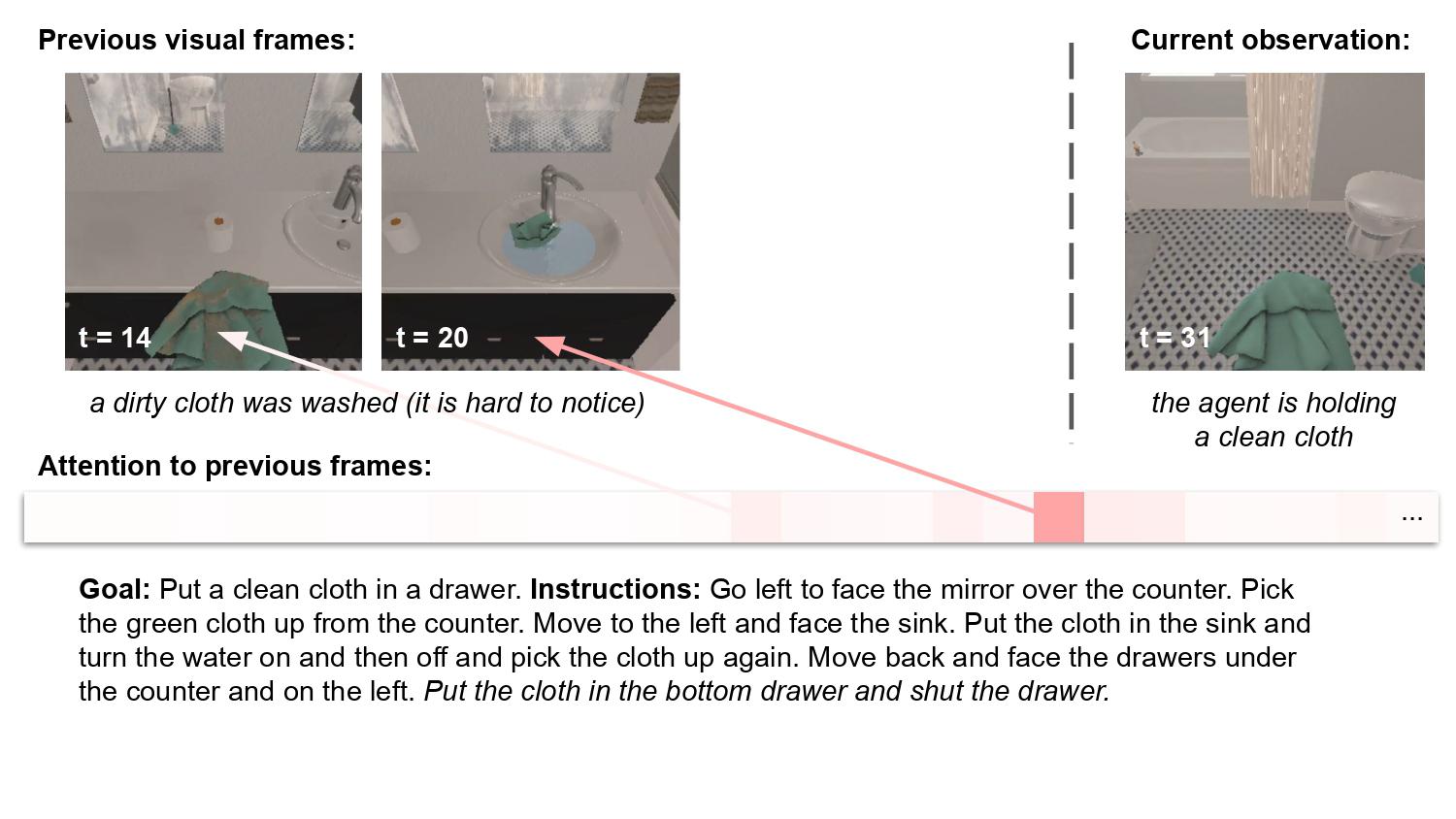}
\vspace{1cm}
\caption{A visualization of normalized attention heatmap to previous visual observations. In this example, the agent is asked to wash a cloth and to put it in a cupboard. The agent washes the cloth at timestep $20$ but the washed cloth does not look very different from a dirty one. At timestep $31$, the agent attends to the previous frames where the washing action is visible to keep track of the cloth state change.}
\label{fig:attn_vis4}
\end{figure*}

To better understand the impact of using a transformer encoder for action predictions, we show several qualitative examples of attention weights produced by the multimodal encoder of an E.T. agent. We use attention rollout~\cite{attentionrollout} to compute attention weights from an output action to previous visual observations. Attention rollout averages attention of all heads and recursively multiplies attention weights of all transformer layers taking into account skip connections.
Figures~\ref{fig:attn_vis1}-\ref{fig:attn_vis4} show examples where an E.T. model attends to previous visual frames to successfully solve a task. The frames attention weights are showed with a horizontal bar where frames corresponding to white squares have close to zero attention scores and frames corresponding to red squares have high attention scores. We do not include the attention score of the current frame as it is always significantly higher than scores for previous frames.

In Figure~\ref{fig:attn_vis1} the agent is asked to pick up an apple and to heat it using a microwave. The agent walks past a microwave at timestep $8$, picks up an apple at timestep $18$ and attends to the microwave frame in order to recall where to bring the apple.
In Figure~\ref{fig:attn_vis2} the agent slices a potato at timesteps $17-18$ (hard to see on the visual observations). Later, the agent gets rid of the knife and follows the next instruction asking to pick up a potato slice. At timestep $39$, the agent attends to the frames $17-18$ where the potato was sliced in order to come back to the slices and complete the task.
In Figure~\ref{fig:attn_vis3} the agent needs to sequentially move two pans. While picking up the second pan at timestep $29$, the agent attends to the frames $20-22$ where the first pan was replaced.
In Figure~\ref{fig:attn_vis4} the agent is asked to wash a cloth and to put it to a drawer. The agent washes the cloth at timestep $20$ but the cloth state change is hard to notice at the given frames. At timestep $31$, the agent attends to the frame with an open tap in order to keep track of the cloth state change.
To sum up, the qualitative analysis of the attention mechanism over previous visual frames shows that they are used by the agent to solve challenging tasks and aligns with the quantitative results presented in Section~\ref{sec:4.3}. 

\subsection{Visualizing language attention}
\label{app:attn_lang}

\begin{figure*}[t!]
\centering
\includegraphics[trim={0 10.5cm 0 0}, width=\linewidth]{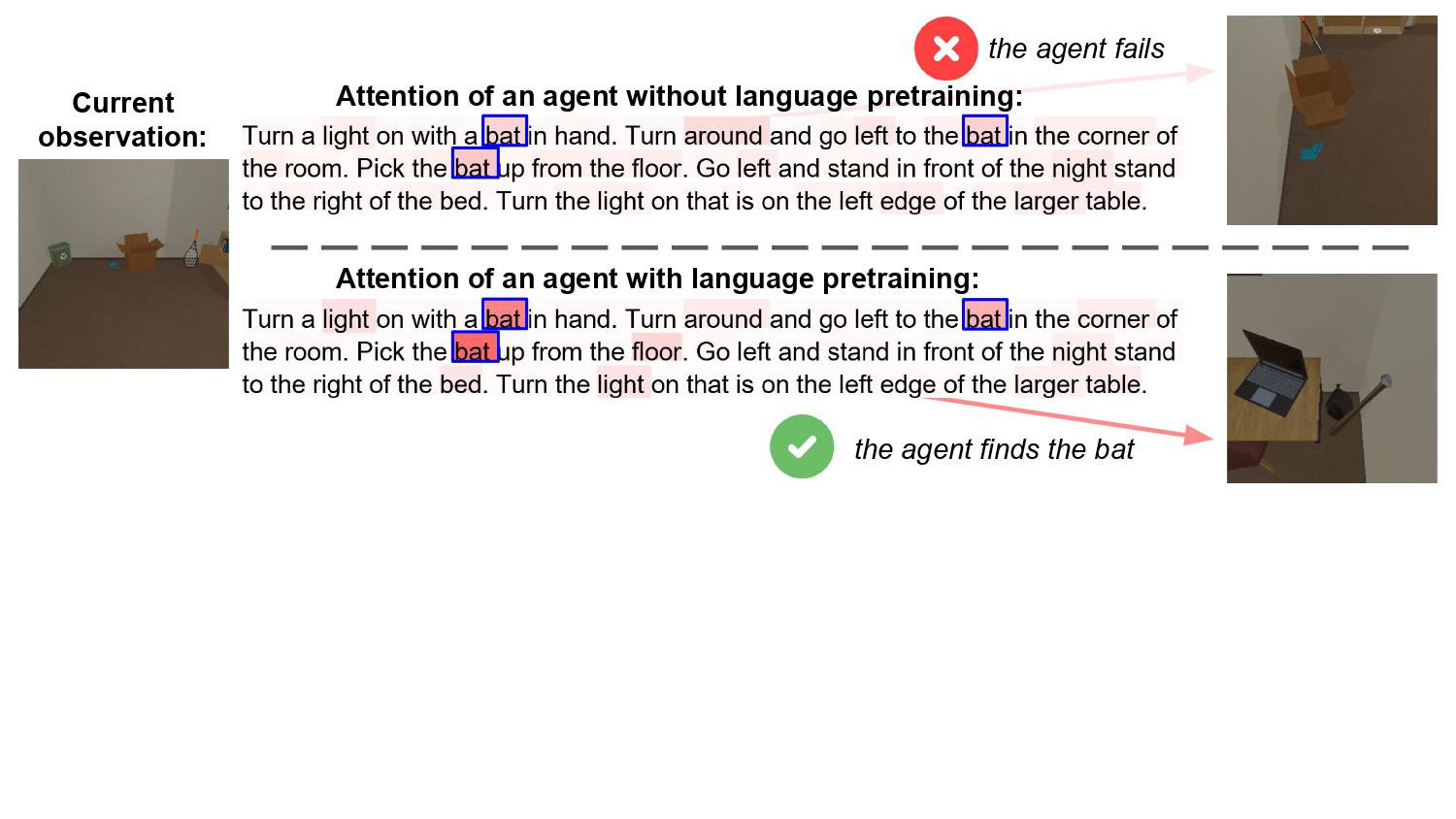}
\includegraphics[trim={0 10cm 0 0}, width=\linewidth]{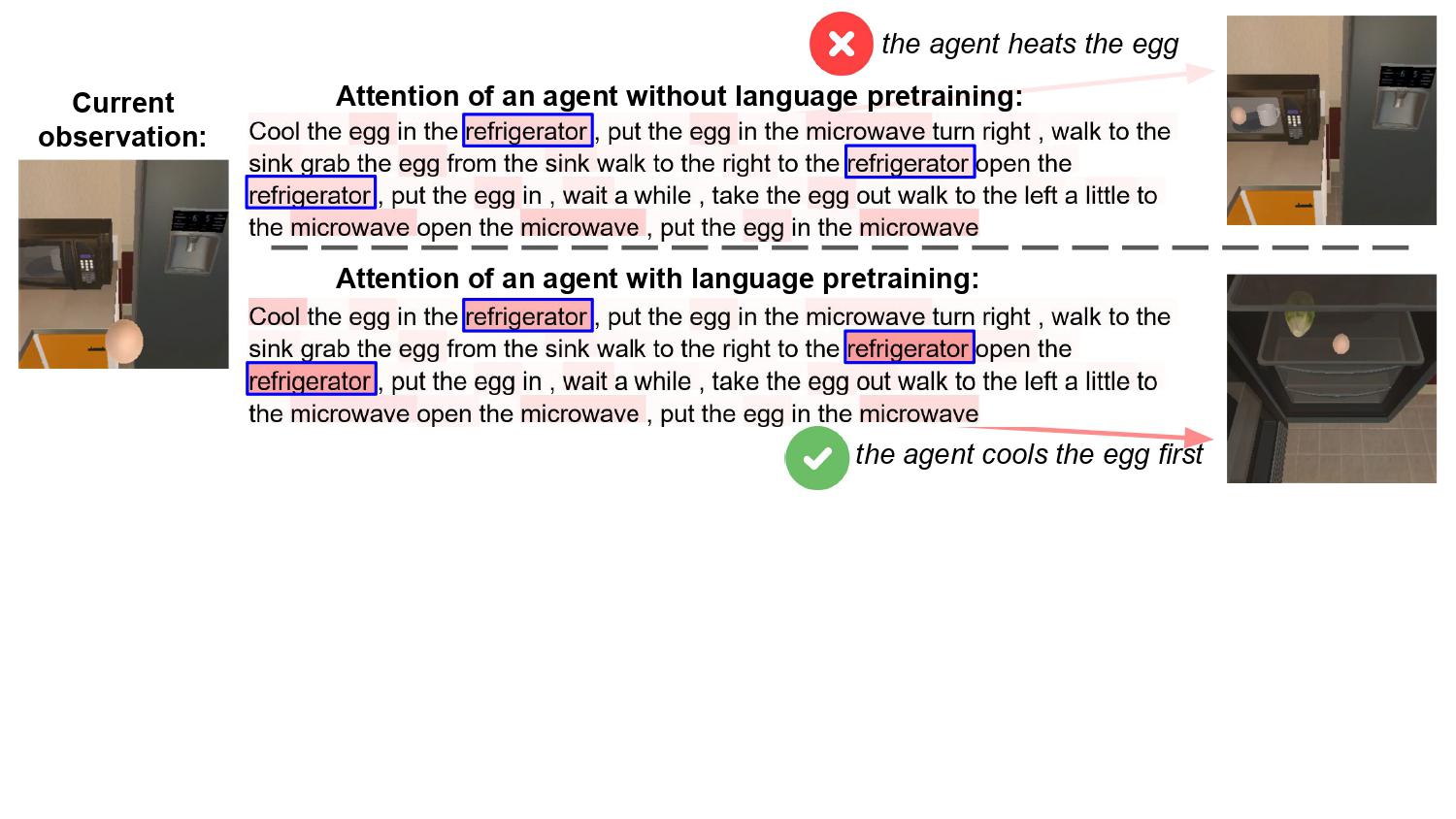}
\includegraphics[trim={0 10cm 0 0}, width=\linewidth]{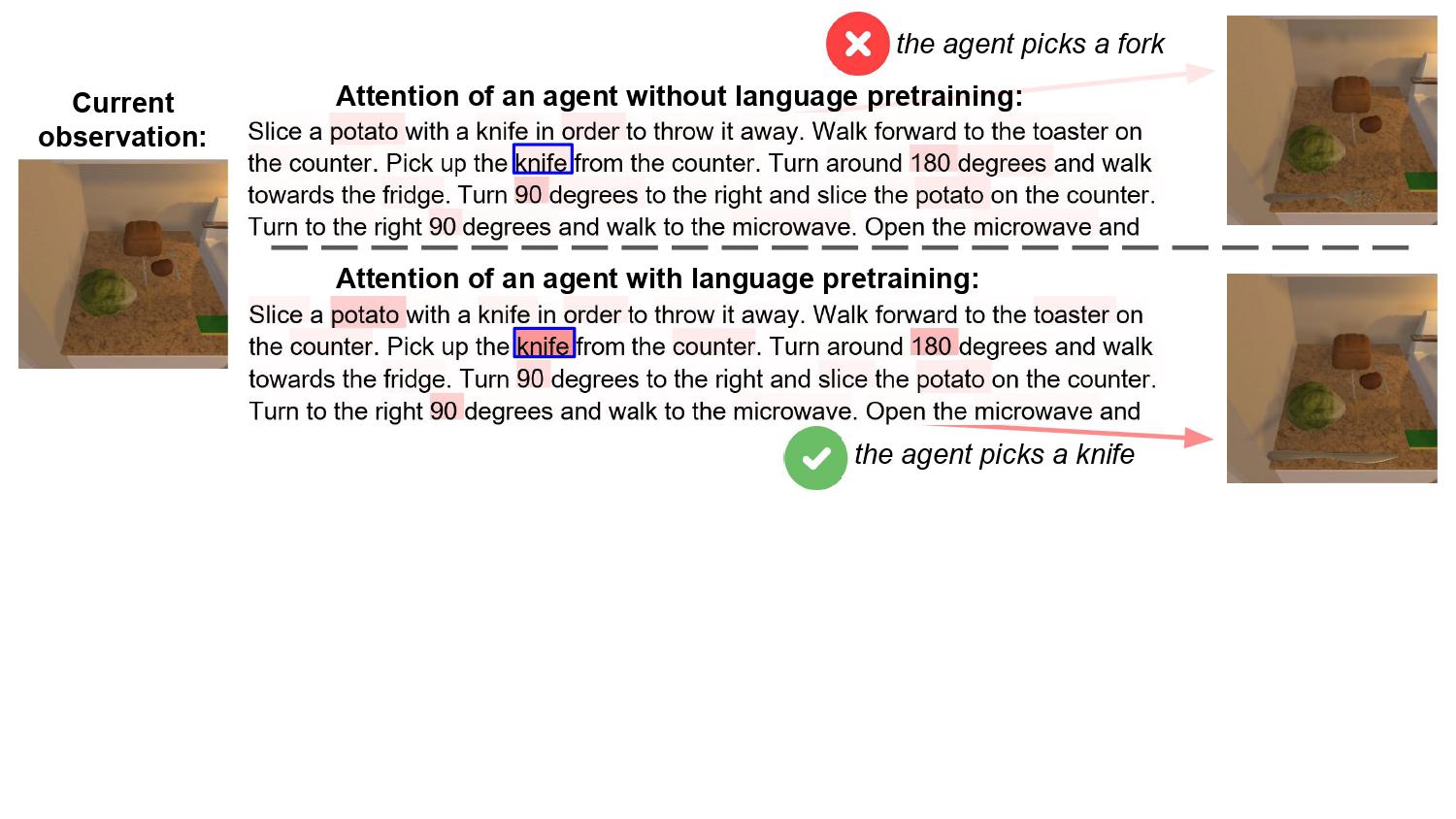}
\caption{Visualizations of normalized language attention heatmaps, without and with the language encoder pretraining. Red indicates a higher attention score. We observe that the agent trained without language pretraining misses word tokens that are important for the task according to human interpretation (marked with blue rectangles). In contrast, the pretrained E.T. agent often is able to pay attention to those tokens and solve the tasks successfully.}
\label{fig:attn_lang}
\end{figure*}

Figure~\ref{fig:attn_lang} illustrates transformer attention scores from an output action to input language tokens by comparing two models: (1) E.T. model trained from scratch, (2) E.T. model whose language encoder is pretrained as in Section~\ref{sec:3.3}. Similarly to the visual  attention, we use attention rollout and highlight the words with high attention scores with red background color.

In the first example of Figure~\ref{fig:attn_lang}, the agent needs to pick up a bat. While the non-pretrained E.T. model has approximately equal attention scores for multiple tokens (those words are highlighted with pale pink color) and does not solve the task, the pretrained E.T. attends to ``bat'' tokens (highlighted
with red) and successfully finds the bat.
In the second example, the agent needs to first cool an egg in a fridge and to heat it in a microwave later. The non-pretrained E.T. has the similar attention scores for ``microwave'' and ``refridgerator'' tokens (they are highlighted with pink) and makes a mistake by choosing to heat the egg first. The pretrained E.T. agent has higher attention scores for the ``refridgerator'' tokens and correctly decides to cool the egg first.
In the third example, the agent needs to pick up a knife to cut a potato later. The non-pretrained agent distributes its attention over many language tokens and picks up a fork which is incorrect. The pretrained E.T. agent strongly attends to the ``knife'' token and picks the knife up.
The demonstrated examples show that the language pretraining of E.T. results in language attention that is better aligned with human interpretation.

\subsection{Qualitative analysis}
\label{app:examples}

\begin{figure*}
    \alframe{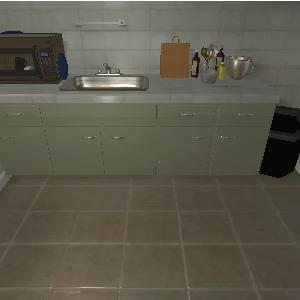}{0}
    \hfill
    \alframe{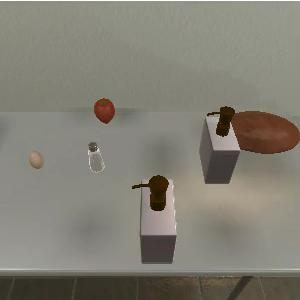}{4}
    \hfill
    \alframe{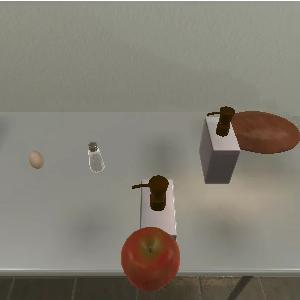}{5}
    \hfill
    \alframe{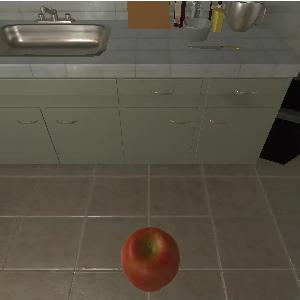}{26}
    \hfill
    \alframe{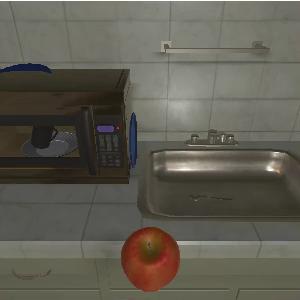}{36}
    \alframe{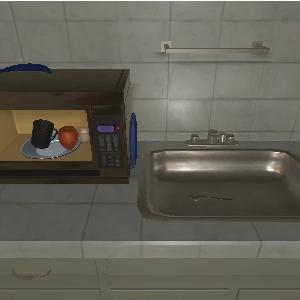}{38}
    \hfill
    \alframe{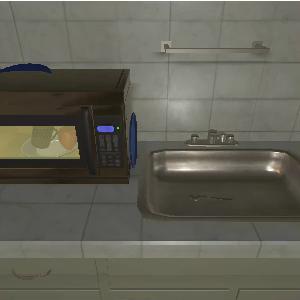}{40}
    \hfill
    \alframe{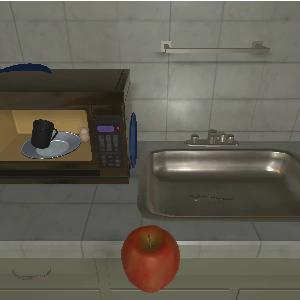}{43}
    \hfill
    \alframe{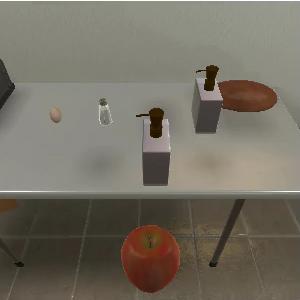}{56}
    \hfill
    \alframe{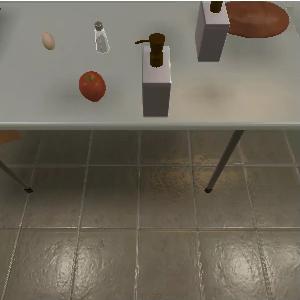}{58} \\
    \textbf{Goal}: Put a heated apple on the table.
    \textbf{Instructions}: Turn left and go to the table.
    Pick up the apple on the table.
    Go right and bring the apple to the microwave.
    Heat the apple in the microwave.
    \textit{Bring the heated apple back to the table on the side.}
    Put the heated apple on the table in front of the salt.
    \caption{Example of a successfully solved task. The agent picks up an apple, puts it into a microwave, closes it, turns it on, opens it, picks up the apple again, then navigates \textit{back to the table on the side} and puts the apple on the same table.
    }
    \label{fig:success1}
\end{figure*}
\begin{figure*}
    \alframe{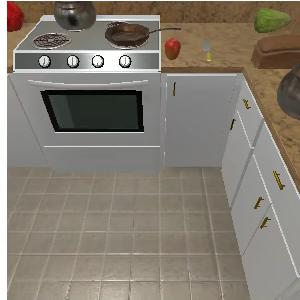}{0}
    \hfill
    \alframe{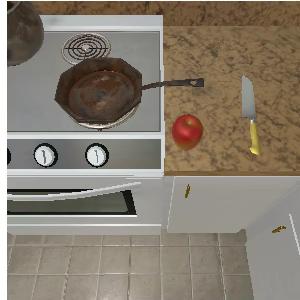}{6}
    \hfill
    \alframe{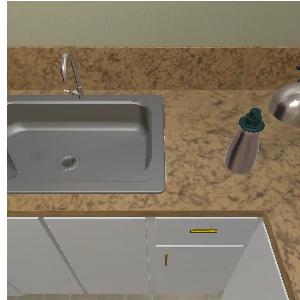}{12}
    \hfill
    \alframe{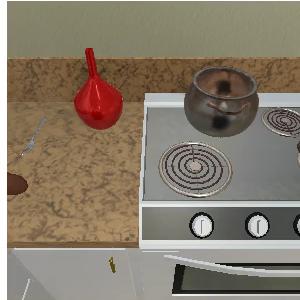}{13}
    \hfill
    \alframe{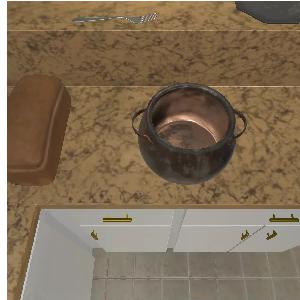}{20}
    \alframe{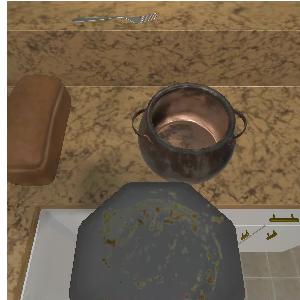}{21}
    \hfill
    \alframe{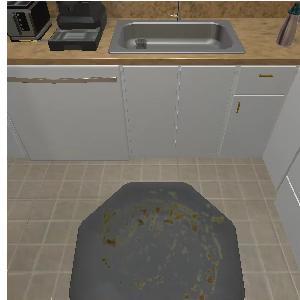}{26}
    \hfill
    \alframe{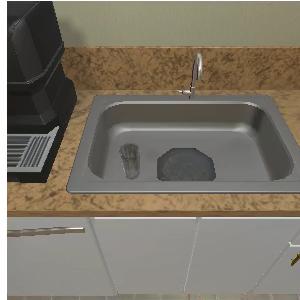}{31}
    \hfill
    \alframe{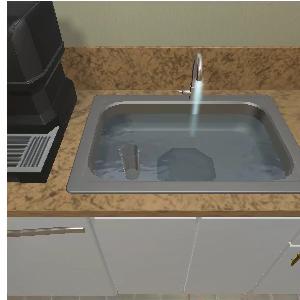}{32}
    \hfill
    \alframe{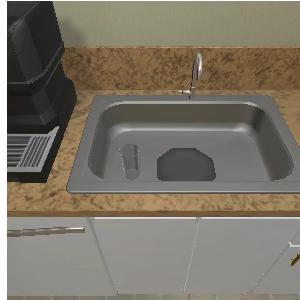}{33}
    \alframe{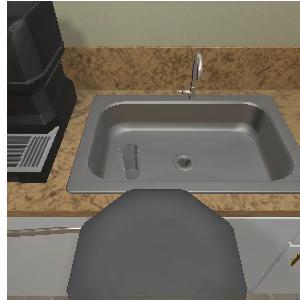}{34}
    \hfill
    \alframe{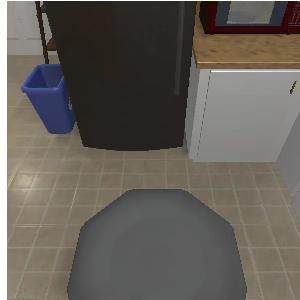}{39}
    \hfill
    \alframe{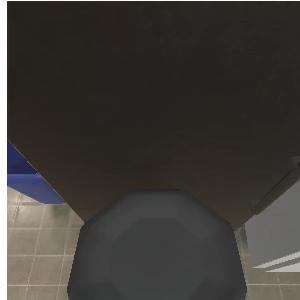}{48}
    \hfill
    \alframe{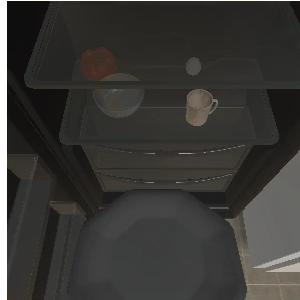}{49}
    \hfill
    \alframe{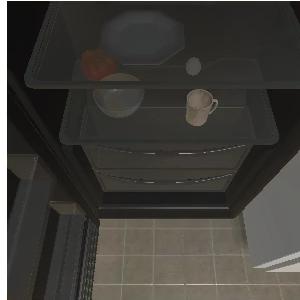}{50} \\
    \textbf{Goal}: Place a rinsed plate in the fridge.
    \textbf{Instructions}:
    Walk ahead to the door, then turn left and take a step, then turn left and face the counter.
    Pick up the dirty plate on the counter.
    Walk left around the counter, and straight to the sink.
    Clean the plate in the sink.
    Turn left and walk to the fridge.
    Place the plate on the top shelf of the fridge.
    Place a pan containing slicing tomato in the refrigerator.
    \caption{Example of a successfully solved task.
    The agent does not know where the dirty plate is and looks at several places on the counter (the first row). It then sees the plate in the corner of the top right image, picks it up, goes to a sink, opens a tap, picks the plate again, navigates to a fridge, opens it and puts the plate to the top shelf of the fridge.
    }
    \label{fig:success3}
\end{figure*}
\begin{figure*}
    \alframe{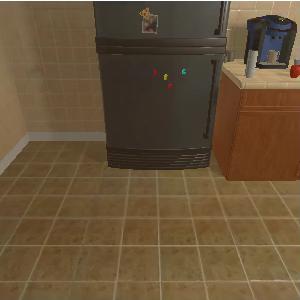}{0}
    \hfill
    \alframe{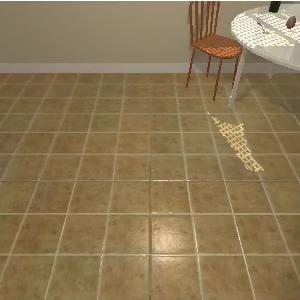}{14}
    \hfill
    \alframe{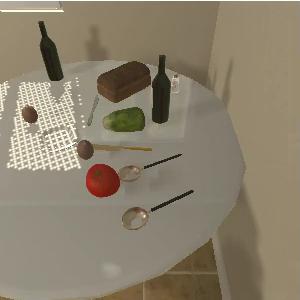}{28}
    \hfill
    \alframe{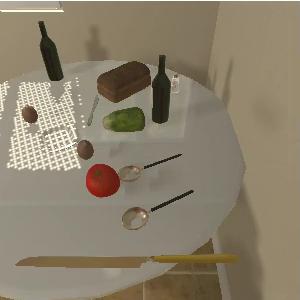}{29}
    \hfill
    \alframe{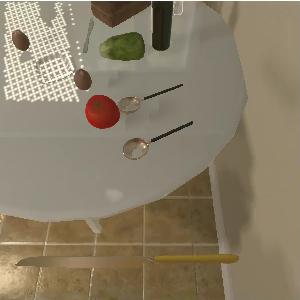}{30}
    \alframe{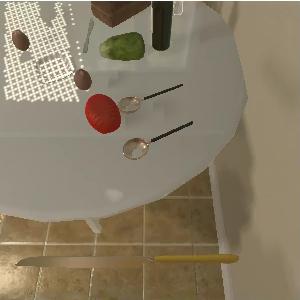}{31}
    \hfill
    \alframe{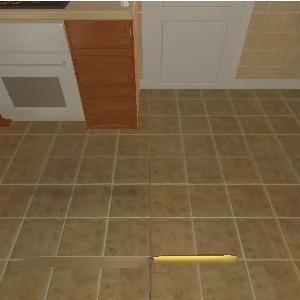}{51}
    \hfill
    \alframe{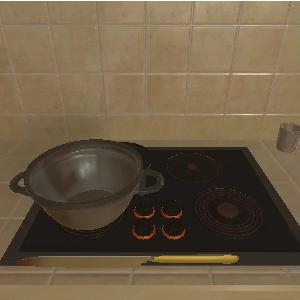}{64}
    \hfill
    \alframe{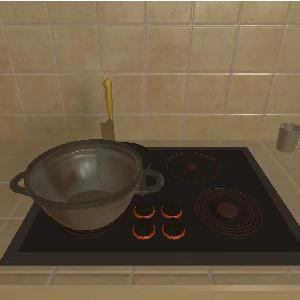}{65}
    \hfill
    \alframe{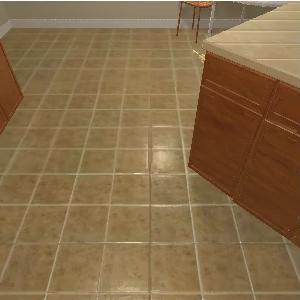}{77}
    \alframe{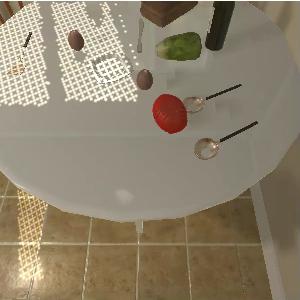}{97}
    \hfill
    \alframe{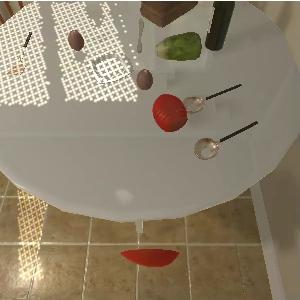}{98}
    \hfill
    \alframe{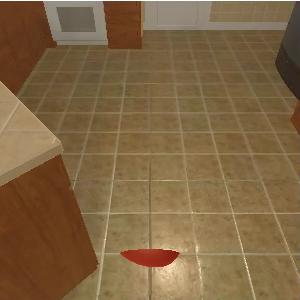}{110}
    \hfill
    \alframe{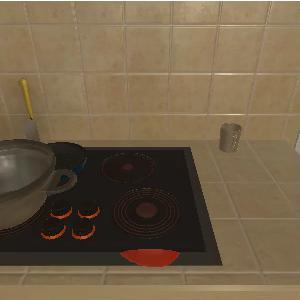}{129}
    \hfill
    \alframe{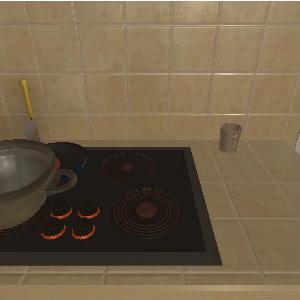}{130}
    \alframe{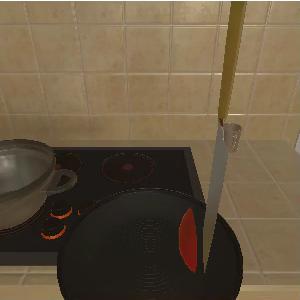}{131}
    \hfill
    \alframe{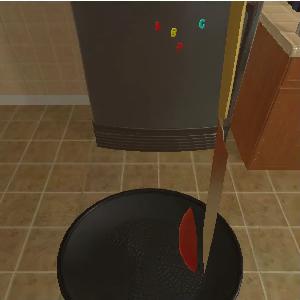}{140}
    \hfill
    \alframe{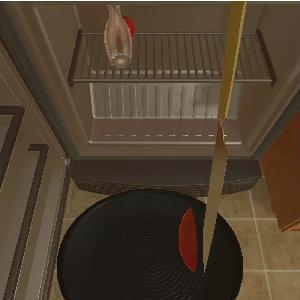}{146}
    \hfill
    \alframe{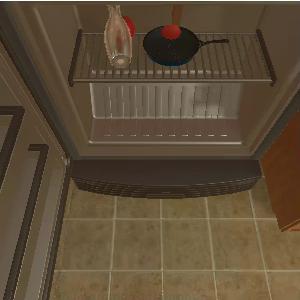}{147}
    \hfill
    \alframe{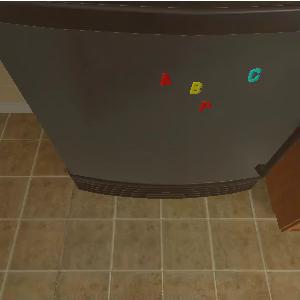}{148} \\
    \textbf{Goal}: Place a pan containing slicing tomato in the refrigerator.
    \textbf{Instructions}: Turn right, move to the table opposite the chair.
    Pick up the knife that is near the tomato.
    Turn left, move to the table opposite the chair.
    Slice the tomato that is on the table.
    Turn left, move to the counter that is left of the bread, right of the potato.
    Put the knife in the pan. Turn left, move to the table opposite the chair.
    Pick up a slice of tomato from the counter.
    Turn left, move to the counter that is left of the bread, right of the potato.
    Put the tomato slice in the pan.
    Pick up the pan from the counter.
    Turn left, move to in front of the refrigerator.
    Put the pan in the refrigerator.
    \caption{Example of a successfully solved task. The agent uses $148$ actions to complete the task. The agent picks up a knife from a table, slices a tomato in the first image of the second row, brings the knife to a stove, puts the knife on a plate, walks back to the table, grabs a tomato slice, returns to the stove, puts the tomato slice on the same plate, picks up the plate, navigates to a fridge, opens it, puts the plate with the knife and the tomato slice on a shelf and closes the fridge.}
    \label{fig:success2}
\end{figure*}
\begin{figure*}
    \alframe{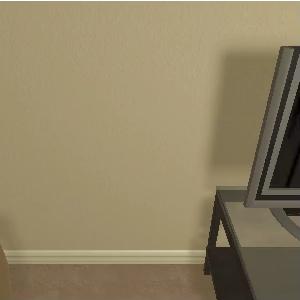}{0}
    \hfill
    \alframe{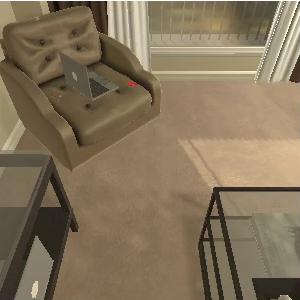}{4}
    \hfill
    \alframe{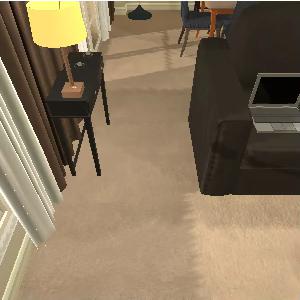}{17}
    \hfill
    \alframe{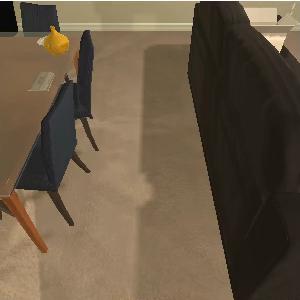}{28}
    \hfill
    \alframe{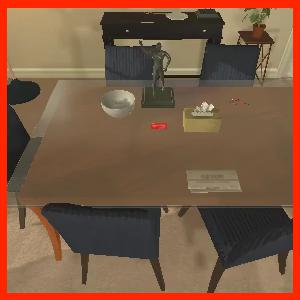}{34}
    \alframe{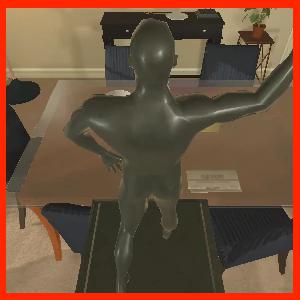}{35}
    \hfill
    \alframe{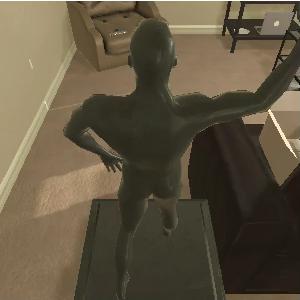}{45}
    \hfill
    \alframe{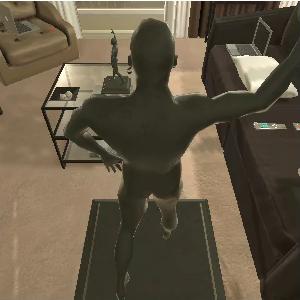}{54}
    \hfill
    \alframe{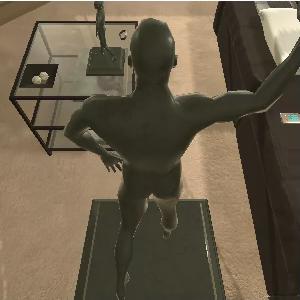}{57}
    \hfill
    \alframe{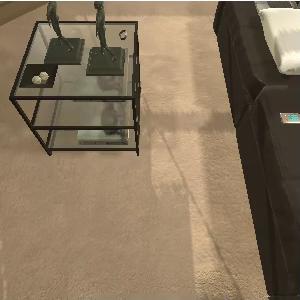}{58}
    \textbf{Goal}: Move a bowl from the table to the coffee table. \textbf{Instructions}: Move across the room to the dining room table where the statue is.
    \textit{Pick up the bowl to the right of the statue on the table.}
    Carry the bowl to the glass coffee table.
    Place the bowl on top of the coffee table between the statue and the square black tray.
    \caption{Failure example in a seen environment. The agent correctly finds both dining and coffee tables but gets confused with \textit{"the bowl to the right of the statue"} reference. The agent decides to pick up a statue instead of a bowl and fails to solve the task.
    }
    \label{fig:failure1}
\end{figure*}
\begin{figure*}
    \alframe{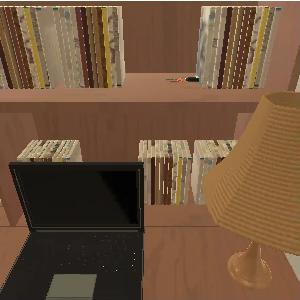}{0}
    \hfill
    \alframe{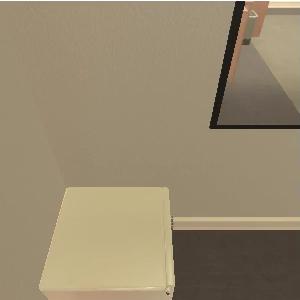}{9}
    \hfill
    \alframe{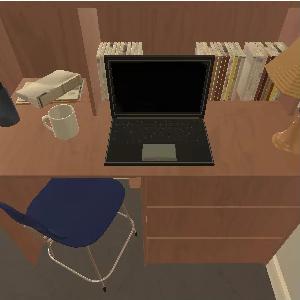}{14}
    \hfill
    \alframe{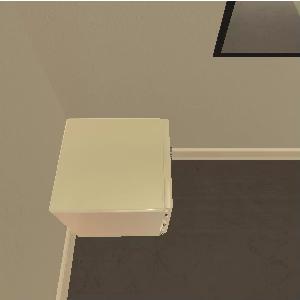}{21}
    \hfill
    \alframe{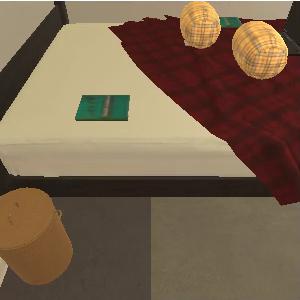}{29}
    \alframe{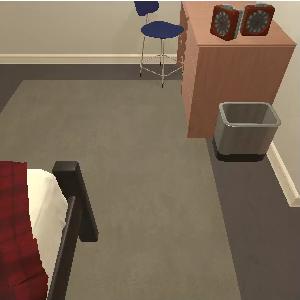}{35}
    \hfill
    \alframe{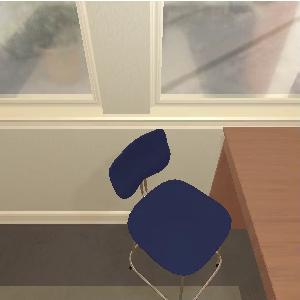}{45}
    \hfill
    \alframe{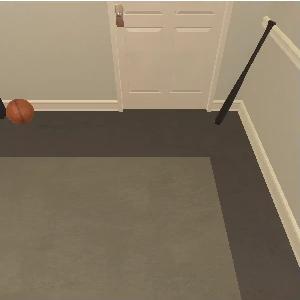}{46}
    \hfill
    \alframe{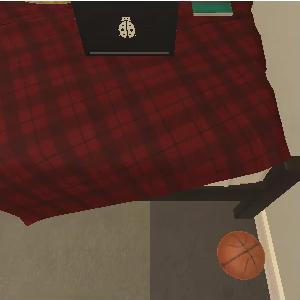}{62}
    \hfill
    \alframe{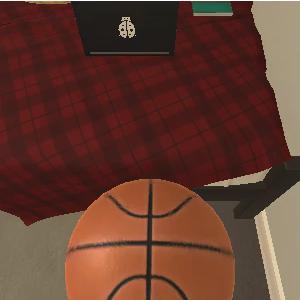}{63}
    \alframe{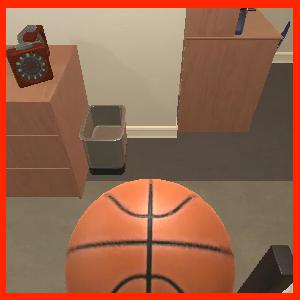}{68}
    \hfill
    \alframe{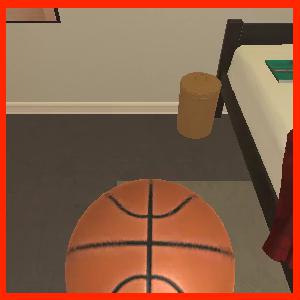}{77}
    \hfill
    \alframe{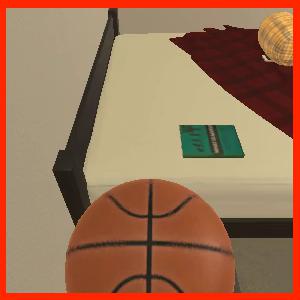}{85}
    \hfill
    \alframe{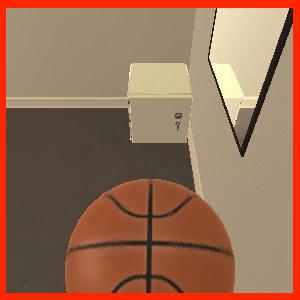}{88}
    \hfill
    \alframe{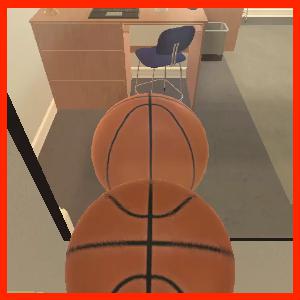}{96}
    \textbf{Goal}: Look at a basketball in the lamp light.
    \textbf{Instructions}: Turn around and go to the foot of the bed.
    Pick up the basketball from the floor.
    \textit{Turn around and go to the desk in the corner.}
    Turn on the lamp.
    \caption{Failure example in an unseen environment. The agent is exposed to an unknown environment and fails to follow the navigation instructions. It wanders around the room, eventually finds a basketball but fails to locate a lamp and decides to terminate the episode in front of a mirror.}
    \label{fig:failure2}
\end{figure*}

We show 3 successful and 2 failed examples of the E.T. agent solving tasks from the ALFRED validation fold.
In Figure~\ref{fig:success1} the agent successfully heats an apple and puts it on a table. The agent understands the instruction \textit{``bring the heated apple back to the table on the side''} and navigates back to its previous position.
In Figure~\ref{fig:success3} the agent brings a washed plate to a fridge. The agent does not know where the plate is and walks along a counter checking several places. Finally, it finds the plate, washes it and brings it to the fridge.
In Figure~\ref{fig:success2} the agent performs a sequence of $148$ actions and successfully solves a task. This example shows that the agent is able to pick up small objects such as a knife and a tomato slice. The agent puts both of them to a plate and brings the plate to a fridge.

Among the most common failure cases are picking up wrong objects and mistakes during navigation.
In Figure~\ref{fig:failure1} the agent misunderstands the instruction \textit{``pick up the bowl to the right of the statue on the table''} and decides to pick up a statue on the frame marked with red. It then brings the statue to a correct location but the full task is considered to be failed.
Figure~\ref{fig:failure2} shows a failure mode in an unseen environment. The agent is asked to pick up a basketball and to bring it to a lamp. The agent first wanders around a room but eventually picks up the basketball. It then fails to locate the lamp and finds itself staring into a mirror. The agent gives up on solving the task and decides to terminate the episode.

\end{document}